\documentclass[manuscript,screen,nonacm]{acmart} %
\usepackage[inline]{enumitem} %
\usepackage{multirow} %
\usepackage{subcaption} %
\usepackage{booktabs}
\usepackage{adjustbox}

\AtBeginDocument{%
  \providecommand\BibTeX{{%
    \normalfont B\kern-0.5em{\scshape i\kern-0.25em b}\kern-0.8em\TeX}}}

\usepackage{tikz}
\usetikzlibrary{shapes.multipart, positioning}
\definecolor{green1}{HTML}{D5E8D4}
\definecolor{orange1}{HTML}{FFE6CC}
\definecolor{blue1}{HTML}{DAE8FC}
\definecolor{purple1}{HTML}{E1D5E7}

\acmConference[Conference acronym 'XX]{Make sure to enter the correct
  conference title from your rights confirmation emai}{June 03--05,
  2018}{Woodstock, NY}
\acmPrice{15.00}
\acmISBN{978-1-4503-XXXX-X/18/06}

\begin{document}

\title[DLAS: Deep Learning Acceleration Stack]{DLAS: An Exploration and Assessment of the \underline{D}eep \underline{L}earning \underline{A}cceleration \underline{S}tack}

\author{Perry Gibson}
\email{p.gibson.2@research.gla.ac.uk}
\orcid{0000-0003-3370-0698}
\author{Jos\'e Cano}
\email{Jose.CanoReyes@glasgow.ac.uk}
\orcid{0000-0002-2243-389X}
\affiliation{%
  \institution{University of Glasgow}
  \city{Glasgow}
  \state{Scotland}
  \country{UK}
  \postcode{G12 8QQ}
}
\author{Elliot J. Crowley}
\email{elliot.j.crowley@ed.ac.uk}
\author{Amos Storkey}
\email{a.storkey@ed.ac.uk}
\author{Michael O'Boyle}
\email{mob@inf.ed.ac.uk}
\affiliation{%
    \institution{University of Edinburgh}
    \city{Edinburgh}
   \state{Scotland}
  \country{UK}
  \postcode{EH8 9YL}
 }
\renewcommand{\shortauthors}{Gibson and Cano, et al.}

\begin{abstract}

Deep Neural Networks (DNNs) are extremely computationally demanding, which presents a large barrier to their deployment on resource-constrained devices.
Since such devices are where many emerging deep learning (DL) applications lie (e.g., obstacle detection for mobile robots, vision-based medical assistive technology), significant bodies of work from both the machine learning and systems communities have attempted to provide optimizations that will make DNNs deployable to edge devices.
To help unify these two perspectives, in this paper we combine machine learning and systems techniques within the Deep Learning Acceleration Stack (DLAS), and demonstrate how these layers can be tightly dependent on each other with an across-stack perturbation study.
We evaluate the impact on accuracy and inference time when varying different parameters of DLAS across two datasets, seven popular DNN architectures, four DNN compression techniques, three algorithmic primitives with sparse and dense variants, untuned and auto-scheduled code generation, and four hardware platforms.
Our evaluation highlights how perturbations across DLAS parameters can cause significant variation and across-stack interactions.
The highest level observation from our evaluation is that the DNN model size, accuracy, and inference time are not guaranteed to be correlated.
Overall we make 13 key observations, including that speedups provided by compression techniques such as quantization are very hardware dependent, and that compiler auto-tuning can significantly alter what the best algorithm to use for a given configuration is.
With DLAS, we aim to provide a reference framework to aid machine learning and systems practitioners in reasoning about the context in which their respective DNN acceleration solutions exist in.
With our evaluation strongly motivating the need for co-design, we believe that DLAS can be a valuable concept for exploring the next generation of co-designed accelerated deep learning solutions.

\end{abstract}

\maketitle

\section{Introduction}
\label{sec:introduction}

Recent years have yielded rapid advances in deep learning, largely due to the unparalleled effectiveness of Deep Neural Networks (DNNs), such as Convolutional Neural Networks (CNNs) and Transformer architectures~\cite{vaswaniAttentionAllYou2017}, on a variety of difficult problems~\cite{lecunDeepLearning2015,jumperHighlyAccurateProtein2021,yamadaLUKEDeepContextualized2020}.
Despite increases in algorithmic efficiency~\cite{hernandezMeasuringAlgorithmicEfficiency2020}, the trend with DNN architectures is increased size and deployment costs as demands for more powerful and general solutions grows~\cite{canzianiAnalysisDeepNeural2017,thompsonComputationalLimitsDeep2022}.
As such, creative approaches are required to deploy DNNs on hardware with limited resources in order to enable a variety of emerging applications (e.g., autonomous driving~\cite{teichmannMultiNetRealtimeJoint2018}, collision avoidance for quadcopters~\cite{alvarezCollisionAvoidanceQuadrotors2016}).
However, often these optimization approaches come with limited benchmarks and few comparisons, and there may be a disconnect between machine learning based and systems based optimizations due to disparate communities with varying core competencies.

Even in a simplified view of the relevant components of deep learning deployment (e.g., machine learning, software, and hardware), it is evident that choices in one area can have consequences for the choices in others~\cite{turnerCharacterisingAcrossStackOptimisations2018}.
For example, constrained CPU hardware will require appropriately resource-conservative software, and DNN models that fit limited memory and latency budgets.
Alternatively, a new DNN architecture with a novel operation will need an optimized software kernel to execute it on hardware that can provide the required inference time.
Without broad awareness of these interactions from practitioners, potential performance may be lost, since novel machine learning techniques may not be fully exploited by systems techniques~\cite{barhamMachineLearningSystems2019,gibsonOptimizingGroupedConvolutions2020}, or machine learning practitioners may not be aware of under-utilized resources available on their hardware platforms.

In this work, we outline a high-level `stack-based' conceptual view of the relevant techniques pertaining to DNN acceleration, and demonstrate how they are linked with a multi-level experimental analysis.
Our goal is to enable a more comprehensive understanding of the performance available under different constraints of inference accuracy, execution time, memory space, and energy consumption.
We introduce the \emph{Deep Learning Acceleration Stack} (DLAS), which provides a map to both machine learning and systems researchers to reason about the impact of their performance optimizations.
We propose DLAS as six layers\footnote{These layers are not to be confused with the computational layers that make up a neural network.}, as shown in Figure~\ref{tab:dlas}, covering parameters more relevant to machine learning (Datasets \& Problem Spaces, Models \& Neural Architectures, and Model Optimizations), and parameters more relevant to systems (Algorithms \& Data Formats, Systems Software, and Hardware).
Each of the layers can be further decomposed into sub-layers, however the intent of DLAS is to provide a starting point for reasoning about across-stack optimization and encourage co-design of accelerated DNN deployments.
The core contributions of this work include:

\begin{itemize}[noitemsep,topsep=0pt]
  \item We introduce the \textit{Deep Learning Acceleration Stack} to reflect the different layers of optimization, from both machine learning and systems, that can be applied to run a DNN model more efficiently on a given target device.

  \item We select parameters to vary at each layer of DLAS (two datasets, four models, three compression techniques, three algorithms, two compilation techniques, four hardware devices), which gives us a vertical slice of DLAS to explore, presenting results on the inference time and accuracy impacts of our variations.

  \item We develop an experimental framework based on Apache TVM~\cite{chenTVMAutomatedEndtoEnd2018} to evaluate our parameters in a consistent environment, extending TVM where required and possible.

  \item We explore across-stack interactions and make 13 key observations from our results, discussing their consequences and highlighting potential improvements that could be made from other works.
\end{itemize}

\begin{figure*}
  \begin{center}
    \begin{adjustbox}{max width=0.98\textwidth}

      \begin{tikzpicture}[
        box/.style={draw, align=center, minimum width=4.1cm, minimum height=0.6cm},
        tbox/.style={align=center}
        ]
\node[box, fill=green1] (box1) at (0, 0) {\small Datasets \& Problem Spaces};

\node[box, fill=green1, below = 1mm of box1] (box2) {\small Models \& Neural Architectures};

\node[box, fill=green1, below = 1mm of box2] (box3) {\small Model Optimizations};

\node[box, fill=blue1, below = 1mm of box3] (box4) {\small Algorithms \& Data Formats};

\node[box, fill=blue1, below = 1mm of box4] (box5) {\small Systems Software};

\node[box, fill=blue1, below = 1mm of box5] (box6) {\small Hardware};

\node[tbox, right=3mm of box1.east] (text1) {\footnotesize ImageNet~\cite{dengImageNetLargescaleHierarchical2009}, CIFAR-10~\cite{cifar10}, GLUE~\cite{wangGLUEMultiTaskBenchmark2023}; protein folding, pose estimation, natural language understanding};

\node[tbox, right=3mm of box2.east] (text2) {\footnotesize CNNs, Transformers~\cite{vaswaniAttentionAllYou2017}, GANs~\cite{goodfellowGenerativeAdversarialNetworks2020}, Diffusion; MobileNets~\cite{howardMobileNetsEfficientConvolutional2017,sandlerMobileNetV2InvertedResiduals2019}, ResNets~\cite{heDeepResidualLearning2016}, BERT~\cite{devlinBERTPretrainingDeep2019}, Stable Diffusion~\cite{rombachHighresolutionImageSynthesis2022}};

\node[tbox, right=3mm of box3.east] (text3) {\footnotesize Pruning (structured, unstructured), data-type quantization (int8, BNNs), grouped conv2d, knowledge distilation~\cite{turnerDistillingPerformanceEnhanced2018}};

\node[tbox, right=3mm of box4.east] (text4) {\footnotesize GEMM/direct/Winograd convolution, NCHW/NHWC data layout, row/column-major, CSR/COO/BSR};
\node[tbox, right=3mm of box5.east] (text5) {\footnotesize DNN frameworks, tensor compilers, programming paradigms; TensorFlow~\cite{abadiTensorFlowSystemLargescale2016}, TVM~\cite{chenTVMAutomatedEndtoEnd2018}, CUDA~\cite{nickollsScalableParallelProgramming2008}, OpenCL~\cite{stoneOpenCLParallelProgramming2010}};

\node[tbox, right=3mm of box6.east] (text6) {\footnotesize CPUs, GPUs, FPGAs, TPUs~\cite{jouppiInDatacenterPerformanceAnalysis2017}, reconfigurable ASICs (e.g., MAERI~\cite{kwonMAERIEnablingFlexible2018}); SIMD-units, cache behavior, tensor cores};

\end{tikzpicture}
\end{adjustbox}
\caption{
Overview of DLAS, split between machine learning and systems techniques, with examples. \label{tab:dlas}
}

\end{center}
\end{figure*}

The rest of the paper is organized as follows:
in Section~\ref{sec:dlas_stack} we motivate and describe DLAS;
Section~\ref{sec:background} gives the necessary background to understand the optimization techniques we explore across DLAS;
For our evaluation, we describe the experimental setup in Section~\ref{sec:setup} and the results are presented in Section~\ref{sec:evaluation};
Finally, in Section~\ref{sec:discussion} we discuss the results of our evaluation as well as related work which could be exploited in further exploration.

\section{Deep Learning Acceleration Stack}
\label{sec:dlas_stack}

\subsection{Motivation}

The recent growth of deep learning has been partially facilitated by the computational power of high-end GPUs, as well as improvements in algorithmic representations~\cite{hernandezMeasuringAlgorithmicEfficiency2020}.
When combined with a tendency to focus narrowly on higher accuracies and the availability of large server-class GPUs, this has led to state-of-the-art DNN models to explode in size~\cite{pattersonCarbonEmissionsLarge2021}.
This presents a large barrier to deploying many modern machine learning applications on constrained devices.

Both machine learning researchers and systems engineers have proposed innovative solutions to overcome this barrier.
However, these solutions are typically developed in isolation, meaning that machine learning practitioners may not explore the systems consequences of their approach, and vice-versa.
For instance, sparsity is regarded by some in the machine learning community as a silver bullet for compressing models, whereas exploiting parallelism is generally seen as essential by system architects.
Challenging these isolated preconceptions reveals that sparsity does not always excel at reducing the number of operations during inference, and parallelism does not necessarily bring the expected speedups.
These observations are presented in greater detail in Section~\ref{sec:discussion}.

The goal of DLAS is to make it clearer to both machine learning and systems practitioners what the relevant contributors to performance for their DNN workloads are, allowing greater opportunities for co-design and co-optimization.
This is not to advocate for machine learning experts to re-train as systems experts and vice-versa.
Rather, we aim to provide a framework of reasoning, so that practitioners can understand the context in which their area of expertise exists in, and give a `checklist' of other relevant performance contributing factors to be aware of.
By exposing the wide range of choices, and highlighting the impact of across-stack interaction, we also hope to encourage better tooling, so that practitioners can more easily experiment with perturbations.

\subsection{Description of the Stack}
\label{subsec:basic_stack_description}

We introduce the \textit{Deep Learning Acceleration Stack} (DLAS), which spans from the machine learning domain all the way down to the hardware domain.
Each layer can be tuned to optimize different goals (i.e., inference accuracy, execution time, memory footprint, power), or to yield further improvements in adjacent layers.
However, for their potential to be fully realized, many optimizations are required to be implemented using techniques across several layers, i.e., co-design and co-optimization.
DLAS contains the following six layers, with examples given in Figure~\ref{tab:dlas}:

\begin{enumerate}
    \item \textit{Datasets \& Problem Spaces}: the top-level of the stack defines the problem and/or environment that the machine learning solution must solve.

    \item \textit{Models \& Neural Architectures}: DNN models and families of architectures, as well as their training techniques.

    \item \textit{Model Optimizations}: approaches to reduce the size and costs of a DNN model (e.g., memory, inference time), while attempting to maintain the accuracy.

    \item \textit{Algorithms \& Data Formats}: DNN layers (e.g., convolutions) can be implemented using various algorithms, with myriad trade-offs in space and time.
    Interlinked with algorithms are data formats, i.e., how data is laid out in memory.
    These choices can be consistent across a DNN model or vary per layer.

    \item \textit{Systems Software}: Software used to run the DNNs, such as DNN frameworks, algorithmic implementations, supporting infrastructure, tensor compilers, and code-generators.

    \item \textit{Hardware}: Devices the DNN is deployed on, from general purpose hardware (e.g., CPUs, GPUs), to application specific accelerators (e.g., FPGAs, NPUs, TPUs).
    It also includes hardware features, such as SIMD-units, cache behavior, and tensor cores.
\end{enumerate}

Although we have delineated the layers of the stack, it is critical to highlight that design decisions made at each layer of DLAS can directly impact adjacent layers.
They can also influence design decisions across the entire stack.
In addition, a given layer may need to be subdivided into sub-layers by a domain expert, and increased co-design may blur the separation between layers.
However, we believe this six layer structure strikes a balance between descriptiveness and simplicity.
In this paper we perform an across-stack perturbation of some parameters from each layer to determine their impact on inference and accuracy performance, and any interactions between parameters.
In the future, practitioners will increasingly need to be aware of these across-stack interactions, as Moore's law scaling can no longer be relied upon by machine learning engineers~\cite{hennessyNewGoldenAge2018}, and increased competition between hardware designers will require progressively more innovative workload-aware approaches.

\subsection{Case Study}

In the remainder of this paper, we demonstrate the value of DLAS with a case study, choosing a small subset of popular parameters at each layer and showing how they can influence each other.
Note that even examining a small number of parameters can result in a large number of experiment variants, due to the combinatorial growth for each new parameter added --- there are nearly 1,000 combinations in our study alone.
Considering a wider range of parameters poses significant research challenges regarding efficient design space exploration (DSE), which is a broader problem the community continues to tackle from a number of directions~\cite{tollenaereAutotuningConvolutionsEasier2023,andersonOptimalDNNPrimitive2018,gibsonTransferTuningReusingAutoSchedules2023,caiOnceforAllTrainOne2020}.

Our case study highlights how an across-stack DLAS evaluation can be conducted, with a narrow but deep investigation.
We encourage readers to consider how the results could change if additional parameters were included, for example Winograd convolution~\cite{lavinFastAlgorithmsConvolutional2016}, Transformers~\cite{vaswaniAttentionAllYou2017}, TPUs~\cite{jouppiInDatacenterPerformanceAnalysis2017}, or some other acceleration technique of interest.
However, including these additional parameters will not change the core purpose of the case study, namely to highlight that DLAS can be a useful delineation for DNN acceleration research, and across-stack thinking will be increasingly important to unlock the next generation of acceleration techniques.
Section~\ref{sec:background} gives the necessary background to understand the details of our case study, Section~\ref{sec:setup} describes the experimental setup, and Sections~\ref{sec:evaluation} and~\ref{sec:discussion} provide the results and discussion respectively.

\section{Background}
\label{sec:background}

In this section we discuss the necessary background of the techniques in DLAS explored in our experiments. Note that providing a complete background of every technique in DLAS is beyond the scope of this paper, as deep learning innovations are rapid and continuous, making any attempt at comprehensiveness quickly outdated.

\subsection{Datasets \& Problem Spaces}
\label{subsec:background:datasets}

We focus on two common image classification datasets, CIFAR-10~\cite{cifar10} and ImageNet~\cite{dengImageNetLargescaleHierarchical2009}, for predicting the class of a square RGB image.
CIFAR-10 includes 60,000 images of 32$\times$32 pixels across 10 classes, while ImageNet has over 14 million images of 224$\times$224 pixels across 1000 classes.

\subsection{Models \& Neural Architectures}

There is a wide-range of DNN architectures available, with a variety of popular models for each.
Convolutional neural networks (CNNs) are commonly leveraged for image classification tasks, and are characterized by their use of convolutional layers.
Other layers may include batch normalization and pooling layers, fully-connected layers, and activation functions such as ReLU.
The topology of these networks are generally deterministic, directed acyclic graphs.
Some neural architectures may include skip-connections~\cite{heDeepResidualLearning2016}, where activations from previous layers are reused in later layers, or depthwise separable convolutions~\cite{sifreRigidmotionScatteringImage2014}, which can reduce memory and computational requirements.

When training DNNs, we repeatedly present the full dataset (each round called an epoch) and assess ultimate performance against a separate test set.
Algorithms like stochastic gradient descent (SGD) iteratively update the DNN weights to enhance accuracy between epochs.
Hyperparameters like the learning rate (LR) influence weight adjustments each epoch, usually reducing as the network learns.

\subsection{Model Optimizations}

\begin{figure*}[t]
\begin{center}
  \begin{subfigure}{0.21\textwidth}
  	\includegraphics[width=\linewidth]{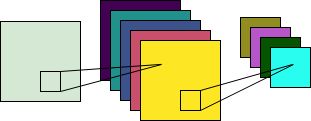}
    \caption{\footnotesize Dense Convolutions}
\label{fig:dense}
  \end{subfigure}
    \hspace*{\fill}   %
  \begin{subfigure}{0.21\textwidth}
  \includegraphics[width=\linewidth]{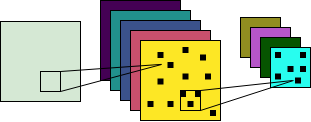}
    \caption{\footnotesize Unstructured Pruning}
  \label{fig:unstructure_pruning}
  \end{subfigure}
  \hspace*{\fill}   %
  \begin{subfigure}{0.21\textwidth}
     \includegraphics[width=\linewidth]{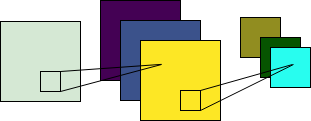}
    \caption{\footnotesize Structured Pruning}
  \label{fig:channel_pruning}
  \end{subfigure}
    \hspace*{\fill}   %
  \begin{subfigure}{0.21\textwidth}
  \includegraphics[width=\linewidth]{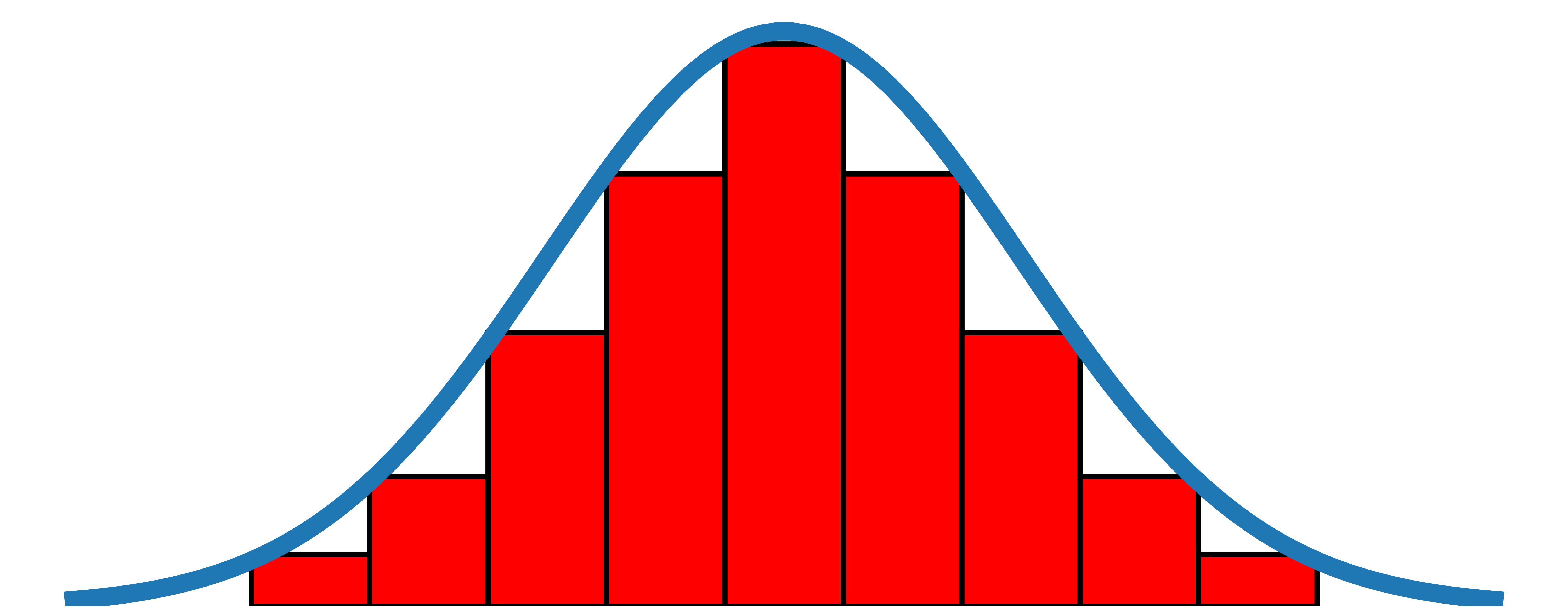}
    \caption{\footnotesize Data-type Quantization}
      \label{fig:quantization}
  \end{subfigure}
\caption{
A visual representation of different Model Optimization techniques: (a) shows a slice of a typical DNN. Input on the left is computed with filters in the center, to produce output on the right; (b) shows the same network slice with weight pruning applied. A subset of parameters in the filters are forced to zero (visually represented with black holes) producing sparse matrices; (c) shows the network slice with channel pruning applied onto the filters, where there are fewer channels; (d) shows data-type quantization, where the range of values that a given parameter of a DNN has been reduced.
}
\label{fig:compression_techniques}
\end{center}
\end{figure*}

A common observation is that DNN models are overparameterized, and similar accuracies can be achieved with smaller models~\cite{frankleLotteryTicketHypothesis2019}.
As a result, a wide range of model optimization and compression techniques have been proposed in the machine learning community.
Figure~\ref{fig:compression_techniques} shows some common compression techniques, with an uncompressed (or `dense') convolution shown in Figure~\ref{fig:dense}.

Pruning is a family of techniques which set weights/parameters in a DNN to zero.
This can reduce the computational or memory overheads of a given DNN model, with potential accuracy loss.
There are two general types of parameter pruning: unstructured and structured, each of which can be used in either global or layer-wise configurations~\cite{blalockWhatStateNeural2020}.
Unstructured pruning removes individual parameters, as seen in Figure~\ref{fig:unstructure_pruning},
whereas structured pruning removes whole groups of parameters, such as blocks, or channels as seen in Figure~\ref{fig:channel_pruning}.
With global pruning, given some pruning target (e.g., 50\% of parameters should be pruned), the pruning algorithm will find the best parameters to prune across the whole DNN model, meaning that some layers will be more or less pruned than others.
For layer-wise pruning, we prune by a pre-defined amount per layer, e.g., 50\% pruning.
All types of pruning apply some \emph{scoring} function to the weights in order to determine which are the least important, and therefore are likely to have the lowest impact on accuracy if removed.
A popular approach is `L1-pruning', where we prune parameters that have the lowest absolute value, i.e., closest to zero.
Other ranking approaches include gradient-based methods and Taylor series expansion.

Another compression technique is \emph{data-type quantization}, which reduces the number of bits used to represent data, visualized in Figure~\ref{fig:quantization}, where a continuous function is approximated with nine distinct values.
Typically, DNNs are trained using \texttt{float32}, however when they are deployed we can reduce the precision.
Common quantized data-types include \texttt{float16} and \texttt{int8}, as well as emerging machine learning specific types such as \texttt{bfloat16}.
For some types of data-type quantization, it may be necessary to add additional operations to the DNN, e.g., in many \texttt{int8} DNNs we need to store the partial-sums of MACs (multiply-accumulate operations) using up to 16 bits, and then rescale back into \texttt{int8}.

Note that for many model optimization techniques, including pruning and data-type quantization, it may be necessary to use some form of post-training fine-tuning or calibration to try and recover some of the lost accuracy.
This can involve retraining the non-pruned parameters of the model in the case of pruning, or adjusting constants used in the rescaling operations in the case of data-type quantization, or some combination of the two.

From a systems perspective, many model optimization techniques do not necessarily provide speedups unless lower levels of the stack adequately support it.
For example, hardware that can compute using data-types with fewer bits or algorithms that exploit the pruning to skip operations.

\subsection{Algorithms \& Data Formats}

The layers of a given DNN model can be implemented in a variety of ways, as long as they still provide the same output.
However, the behavior of a given implementation is influenced by the size and shape of the data, the properties of the hardware we are running on, as well as the choices around how we exploit model optimization techniques.

For the CNN models we evaluate in our work, the most important component to optimize is the convolutional layers, since they are generally the most compute and memory intensive.
Algorithms used in this work implementing convolutional layers include \emph{direct}, \emph{GEMM}, and \emph{spatial pack} convolution.
\emph{Direct} convolution applies the convolution in a manner similar to the textbook definition `sliding-window', and does not reshape input data and weights.
\emph{GEMM} convolution reshapes input data into a 2D array, potentially replicating elements using a reshaping algorithm known as `im2col'.
This means that the convolution can be computed as a matrix multiplication.
\emph{Spatial pack} convolution reshapes both in input data and weights, although the weights can be reshaped offline, with data packed into tiles that are ideally loaded once.
Unlike GEMM convolution, the size of the reshaped input data is the same size, and tiling on inputs and weights is intended to exploit data reuse and SIMD vectorization.

Another important component of how we format data is the layout, which is how we order the data in memory, since memory may have a different structure to a given tensor (e.g., 1D memory, but 4D tensors).
For 2D arrays, common formats include `row-major' and `column-major' order, with the former meaning that data in the same row is contiguous in memory and the latter meaning data in the same column is contiguous in memory.
Similarly, for 4D data, which is more relevant to our image classification CNNs, two common formats are \texttt{NCHW} and \texttt{NHWC}, with \texttt{N} representing the batch size and \texttt{C}, \texttt{H}, and \texttt{{W}} representing the number of input channels, the input height, and input width respectively.

To achieve savings from pruning, the Algorithms \& Data Formats layer is critical, since the chosen algorithm must support `sparsity', i.e., exploiting the zeros generated by pruning to skip computation or reduce memory usage.
The computational savings are enabled by the fact that regardless of the value of an input, a multiplication by zero will have no impact on the final output; and the memory savings come from representing sequences of zeros in a more compressed format.
In this work, we use the popular CSR (compressed sparse row) format, which represents 2D data using 3 arrays:
\begin{enumerate*}
    \item The non-zero elements of the parameters (\emph{data});
    \item the original column index of the corresponding parameters (\emph{indices}); and
    \item the first non-zero elements in each row, as well as the final non-zero element (\emph{indptr}).
\end{enumerate*}
Other formats include BSR (block sparse row) and COO (coordinate list).

\subsection{Systems Software}

The systems software most commonly exposed to machine learning practitioners is the DNN framework.
Common frameworks include PyTorch~\cite{paszkePyTorchImperativeStyle2019}, TensorFlow~\cite{abadiTensorFlowSystemLargescale2016}, JAX~\cite{frostigCompilingMachineLearning2018}, and MXNet~\cite{chenMXNetFlexibleEfficient2015}, all of which are focused on DNN training.
Other frameworks may focus exclusively on deployment, such as TensorFlow Lite and TensorRT~\cite{nvidiacorporationNVIDIATensorRTProgrammable2016}.
These frameworks include utilities for defining, saving, and loading models; passing, processing, and inspecting data; and invoking and profiling training and inference.
Underlying these DNN frameworks are the kernel libraries which execute the critical computations of the DNNs (e.g., the convolutional layers).
For example, for Nvidia GPUs, many frameworks leverage the cuDNN library~\cite{chetlurCuDNNEfficientPrimitives2014}, which is a collection of optimized CUDA kernels to run common DNN operations.

An alternative to using optimized vendor libraries is using a tensor compiler such as TVM~\cite{chenTVMAutomatedEndtoEnd2018} or IREE~\cite{theireeauthorsIREE2019}.
They generate code for a specific DNN and hardware backend, and when leveraged correctly can outperform vendor libraries, especially for operations that may be less popular or optimized.
TVM uses a `compute schedule' programming paradigm, similar to Halide~\cite{ragan-kelleyHalideDecouplingAlgorithms2017}, where a high level description of the computation is complemented by `schedules'.
Schedules are platform-specific transformations that are applied to the code for each algorithm to enable better performance.
Examples of schedule language primitives include parallelism, loop unrolling/splitting/reordering, and vectorization.

Tensor compilers can be taken even further, by tuning the code (i.e., schedule) for each layer, using tools such as AutoTVM~\cite{chenLearningOptimizeTensor2018} and  Ansor~\cite{zhengAnsorGeneratingHighPerformance2020}.
Specifically, Ansor is an auto-scheduling system built on top of TVM which automatically searches for optimized schedules for a given DNN model on a given hardware platform.
It leverages a genetic algorithm and learned cost model to iteratively explore schedule transformations.

Below the level of tensor compilers are programming paradigms and general purpose compilers.
LLVM~\cite{Lattner:MSThesis02} is a cross-platform compiler infrastructure which higher level compilers such as TVM can lower their optimized code to, which is then converted to a binary.
For hardware accelerators, systems such as CUDA~\cite{nickollsScalableParallelProgramming2008} and OpenCL~\cite{stoneOpenCLParallelProgramming2010} provide a programming interface for GPUs, and more specialized accelerators may define their own libraries.
Generally, when using accelerators, we still require CPU-side host code to manage accelerator calls and data transfers.

\subsection{Hardware}

Hardware devices that DNN models commonly execute on include CPUs and GPUs, as well as more specialized accelerators.
CPUs are generally complex independent processing cores, typically with one to several dozen cores on a single chip.
GPUs are generally simpler interdependent processing cores, typically with dozens to several thousand cores on a single chip.
Note that this comparison is a simplification, since these cores are not equivalent, and they may vary in speed, programmability, and other features.

Vector or SIMD instructions allow multiple data to be loaded or computed upon using a single instruction.
For example, the Intel instruction \texttt{MOVAPS}
loads four \texttt{float32} values in a single instruction, which in theory represents a 4$\times$ speedup compared to loading them one-by-one.
Practically, micro-architectural considerations means that this speedup may vary.
Different architectures may support varying maximum SIMD length, e.g., Intel's AVX instructions support up to 256 bits (with AVX-512 supporting up to 512 bits), whereas Arm Neon supports up to 128 bits.
Different hardware may also have varying support and levels of optimization for different data-types, e.g., a CPU may support \texttt{float16} data-types but actually execute them as \texttt{float32} instructions.
Whereas a GPU may have explicit \texttt{float16} instructions that can be exploited.

Another important aspect of hardware is the memory system, with fast, small caches being close to computation, and larger, slower memories higher up the hierarchy.
As well as memory management within processor hardware, data transfers between CPU and an accelerator can also be a critical bottleneck to efficient processing.

\section{Experimental Setup}
\label{sec:setup}

Our experiments represent a vertical slice of DLAS, which demonstrates the design choices available and interactions that occur.
Therefore, we do not optimize for every technique that may influence a given result.
Additional techniques from the literature which could be used to further optimize performance, as well as discussion of the trends in our results, are given in Section~\ref{sec:discussion}.
In addition, we develop our experimental framework using Apache TVM, a state-of-the-art tensor compiler.
We extend TVM where necessary and possible, and will make the code available to the community.

\subsection{Models \& Neural Architectures}
\label{subsec:setup:models}

As highlighted in Section~\ref{subsec:background:datasets}, we investigate two image classification datasets: CIFAR-10 and ImageNet.
For CIFAR-10, we use model definitions from a PyTorch-based library~\cite{kuangliuPyTorchLightningCIFAR102023}, which we train from scratch.
We consider four architectures: ResNet18~\cite{heDeepResidualLearning2016}, MobileNetV1~\cite{howardMobileNetsEfficientConvolutional2017}, MobileNetV2~\cite{sandlerMobileNetV2InvertedResiduals2019}, and VGG-16~\cite{simonyanVeryDeepConvolutional2014}.
ResNet18 and VGG-16 are larger models, and MobileNets V1 and V2 are designed to be more resource efficient.
To train the models we used SGD to minimize the cross-entropy loss (averaged across all data items), which penalizes the network for making incorrect classifications.
We used a 1cycle LR scheduler~\cite{smithSuperConvergenceVeryFast2018} with momentum $0.9$, weight decay $5\times10^{-4}$, and an initial LR of $5\times10^{-2}$, trained for 200 epochs.

For our ImageNet models, we use pre-trained models from the TorchVision repository~\cite{maintainersTorchVisionPyTorchComputer2016}.
We consider four architectures: DenseNet161~\cite{huangDenselyConnectedConvolutional2017}, EfficientNetB0~\cite{tanEfficientNetRethinkingModel2019}, ResNet50~\cite{heDeepResidualLearning2016}, and MobileNetV2~\cite{sandlerMobileNetV2InvertedResiduals2019}.
ResNet50 and DenseNet161 are larger models, and MobileNetV2 and EfficientNetB0 are designed to be more resource efficient.
These models are pre-trained with the training configurations described in the TorchVision documentation.

\subsection{Model Optimizations}

We explore three approaches to compression:
\begin{enumerate*}
\item global L1 unstructured pruning, which we call `weight pruning' (Figure~\ref{fig:unstructure_pruning});
\item global L1 structured pruning over convolutional channels, which we call `channel pruning' (Figure~\ref{fig:channel_pruning}); and
\item data-type quantization, exploring \texttt{float16} and \texttt{int8} quantization (Figure~\ref{fig:quantization}).
\end{enumerate*}
Our pruning techniques explore the impact of increasingly higher compression ratios, thus for the evaluation of inference time and for each model and pruning technique, we select the pruning level which has the highest compression ratio before a significant accuracy drop (elbow point) occurs.
To implement our pruning, we leverage PyTorch Lightning~\cite{falconPyTorchLightning2019}, a wrapper of PyTorch which simplifies pruning.
We apply pruning iteratively, starting with the pre-trained unpruned dense models.
To reduce accuracy loss, at each pruning step we apply fine-tuning.
For weight pruning, we start by pruning at 50\%, then increase in step sizes of 10\%, additionally pruning at 95\%, and 99\%.
For channel pruning, we start by pruning at 5\%, then increase in step sizes of 5\%, additionally pruning at 99\%.
In total, each pruning technique gets the same number of fine-tuning epochs shared evenly in each pruning step.
However, we perform channel pruning in a more fine-grained way to compensate for its more coarse-grained approach to removing weights.
For our CIFAR-10 models we use:
210 epochs of fine-tuning;
an initial LR of $5\times10^{-2}$; and
SGD with momentum 0.9, a weight decay $5\times10^{-4}$, and the 1cycle LR scheduler~\cite{smithSuperConvergenceVeryFast2018}.
For our ImageNet models we use:
140 epochs of fine-tuning;
an initial LR of $1\times10^{-3}$; and
SGD with momentum 0.9, weight decay $5\times10^{-4}$, and the cosine annealing LR scheduler~\cite{loshchilovSGDRStochasticGradient2017}.

For data-type quantization, we use TVM's native conversion tool.
To recover the accuracy for \texttt{int8}, we use the ONNXRuntime's~\cite{developersONNXRuntime2018} post-training quantization tool and the validation dataset.
For \texttt{float16}, as we show later in our results, there is no accuracy loss, thus we do not perform any additional steps to recover lost accuracy.

\subsection{Algorithms \& Data Formats}

We evaluate three algorithmic primitives for the convolutional layers:
\begin{enumerate*}
    \item \emph{direct},
    \item \emph{GEMM}, and
    \item \emph{spatial pack} convolution.
\end{enumerate*}
We use both dense and sparse versions of these algorithms, which we implement or extend within TVM v0.8.0.
The same high level algorithm implementation is used for both CPU and GPU.
All of our algorithms use the \texttt{NCHW} data layout, and for both weight and channel pruning we use the CSR sparse data format.

\subsection{Systems Software}
\label{subsec:setup:sys_software}

For each convolution algorithm we implement a minimal TVM schedule that uses thread parallelism.
However, since TVM's performance comes from having more optimized schedules, unoptimized algorithms may give an unrealistic indication of the best algorithm in each setting.
Thus, rather than hand optimize the schedules, and risk introducing bias from inconsistent levels of optimization, we also leverage the Ansor auto-scheduler~\cite{zhengAnsorGeneratingHighPerformance2020} to generate optimized schedules for each DNN layer and algorithm.
For CPU code, we use TVM's LLVM backend with AVX and Neon extensions for our Intel and Arm based CPUs respectively.
For GPU code, we generate OpenCL and CUDA kernels for our Arm and Nvidia GPUs respectively.

For our auto-scheduling (or `tuned') experiments, we allow Ansor to explore up-to 20,000 program variants, with early stopping permitted if no speedups have been observed after 1,000 variants.
Auto-scheduling sparse computations is not fully supported by TVM.
Thus, we employ an approach in TVM called `sparse sketch rules', where we describe a starting point for the auto-scheduler to begin schedule generation.
This works for the CPU, however TVM is unable to support auto-scheduled sparse computations on GPUs in the versions of TVM we have evaluated.
This is because the auto-scheduler has two conflicting requirements:
\begin{enumerate*}
    \item cross-thread reduction, requiring partial sums that must be computed across GPU threads simultaneously; and
    \item loops parallelized over threads must request a static number of threads.
\end{enumerate*}
Both of these conditions cannot be satisfied, since the size of our reduction loop for our algorithms varies depending on how many non-sparse elements there are in a given portion of the computation.
Thus, we cannot tune pruned models on the GPU in our evaluation, which highlights a type of barrier faced by across-stack acceleration researchers, i.e., limited software support for a given combination of DLAS parameters.

\subsection{Hardware}

Table~\ref{tab:hw_platforms} shows the hardware platforms used in our experiments.
For CPU experiments, we use an Intel i7 workstation machine and the HiKey 970 development board.
The i7 has 6 cores, but due to hyper-threading 12 threads are exposed.
By default, TVM uses one thread per core, a default we follow in our experiments.
The HiKey board has an Arm big.LITTLE architecture, meaning that it has 4 more powerful cores (A73@2.4GHz) and 4 less powerful cores (A53@1.8GHz).
For our experiments, we use only the A73 (big) cores, which is the default for TVM.
In principle, with appropriately configured load balancing between cores, using all cores could bring a performance improvement.
However, this is outside the scope of this work, and as we will discuss in Section~\ref{subsec:discuss:hardware}, exposes further across-stack considerations.
For our GPU experiments, we leverage the GPU cores of the HiKey 970 and an Nvidia AGX Xavier devices.

\begin{table*}[t]
\caption{Hardware features of the devices used in the experiments.
For the HiKey 970, we only use the A73 CPU.
}
\begin{center}
\footnotesize
\begin{tabular}{cccccccc}
\toprule
\textbf{Device} & \textbf{CPU} & \textbf{L1 Cache (I+D)} & \textbf{L2 (+L3) Cache} & \textbf{RAM} & \textbf{GPU} & \textbf{GPU API}\\
\midrule
Intel i7 & Intel i7-8700 (6 cores) @ 3.2 GHz & 192K + 192K & 1.5M (+12M) & 16GB DDR3 & - & - \\
\midrule
\multirow{2}{*}{HiKey 970} & \textit{Arm Cortex-A73 (4 cores) @ 2.4 GHz} & \textit{256K + 256K} & \textit{2M shared} & \multirow{2}{*}{6GB LPDDR4} & Mali-G72 & \multirow{2}{*}{OpenCL} \\
 & Arm Cortex-A53 (4 cores) @ 1.8 GHz & 128K + 128K & 1M shared & & (12 cores) &  \\
\midrule
\multirow{2}{*}{AGX Xavier} & \multirow{2}{*}{Arm v8.2 Carmel (4 cores) @ 2.19 GHz} & \multirow{2}{*}{64K + 128K} & \multirow{2}{*}{2M (+ 4M)} & \multirow{2}{*}{16GB LPDDR4x} & Volta & \multirow{2}{*}{CUDA} \\
&&&&&(512 cores) &\\
\bottomrule
\end{tabular}
\label{tab:hw_platforms}
\end{center}
\end{table*}

\subsection{Evaluation Methodology}

On both CPU and GPU experiments, we ensure that devices are single-tenant and repeat experiments to mitigate potential interference from background processes.
We run each experiment 150 times using TVM's \texttt{time\_evaluator} function with a single input image (i.e., batch size 1).
For auto-scheduling, we run our search across 20,000 program variants once per experiment and evaluate the optimized binary 150 times.
We report the median inference time, disregarding an initial warm-up run.

\section{Evaluation}
\label{sec:evaluation}

We split the results between CIFAR-10 (Section~\ref{subsec:eval:cifar10}) and ImageNet (Section~\ref{subsec:eval:imagenet}).
We first analyze the accuracy impact of the optimization techniques and choose maximally compressed models for each technique which maintain accuracy.
Then, we analyze the inference performance of these models using the algorithms and compilation configurations on the CPUs and GPUs.
Section~\ref{sec:discussion} gives a high-level discussion of the results.

\begin{figure*}[t]
  \centering
\begin{subfigure}{0.31\textwidth}
  \centering
  \includegraphics[width=\linewidth]{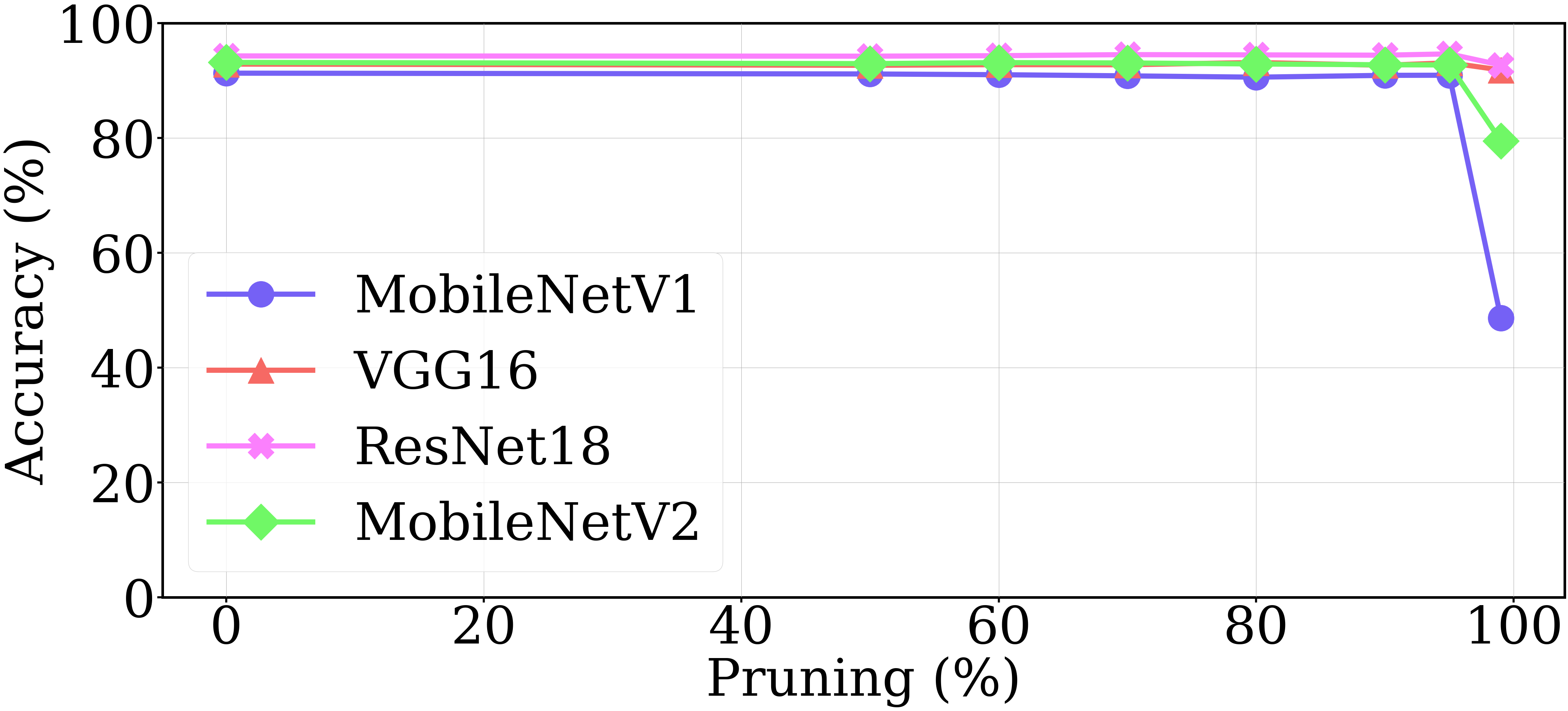}
  \caption{\footnotesize Weight pruning (CIFAR-10)}
  \label{fig:acc_unstructured_cifar10}
 \end{subfigure}
 \hspace{1em}%
\begin{subfigure}{0.31\textwidth}
  \centering
  \includegraphics[width=\linewidth]{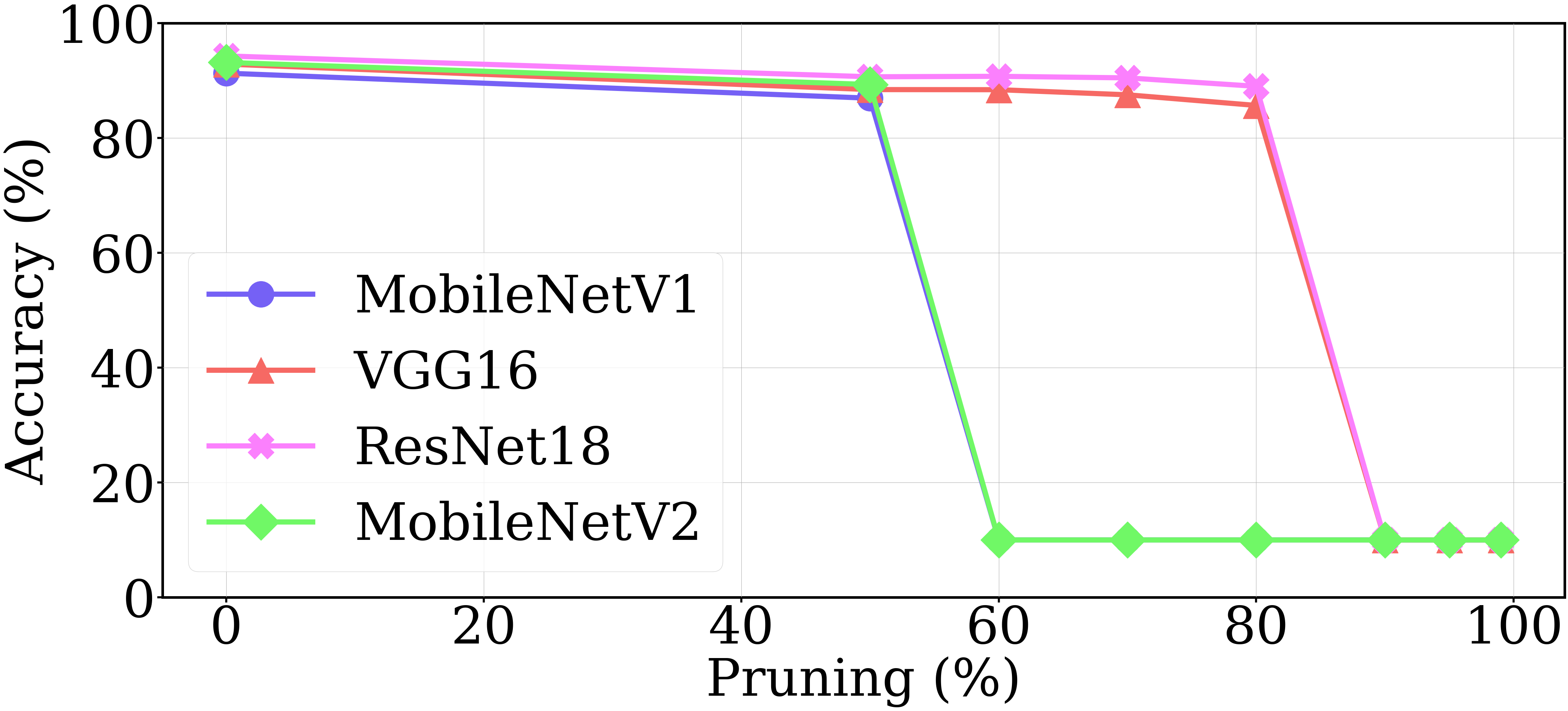}
  \caption{\footnotesize Channel pruning (CIFAR-10)}
  \label{fig:acc_structured_cifar10}
\end{subfigure}
 \hspace{1em}%
\begin{subfigure}{0.31\textwidth}
  \centering
\includegraphics[width=\linewidth]{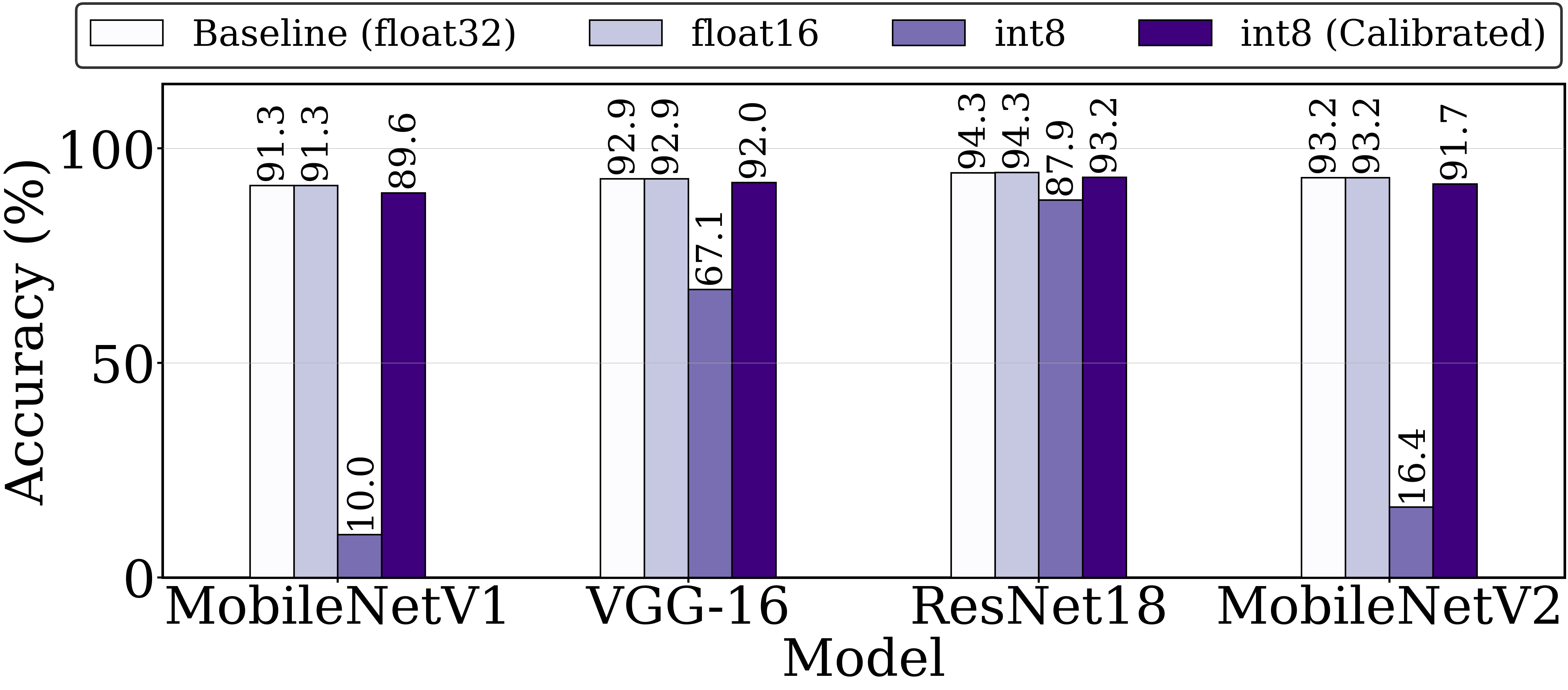}
  \caption{\footnotesize Data-type quantization (CIFAR-10)}
  \label{fig:acc_quant_cifar10}
\end{subfigure} \\

\begin{subfigure}{0.31\textwidth}
  \centering
  \includegraphics[width=\linewidth]{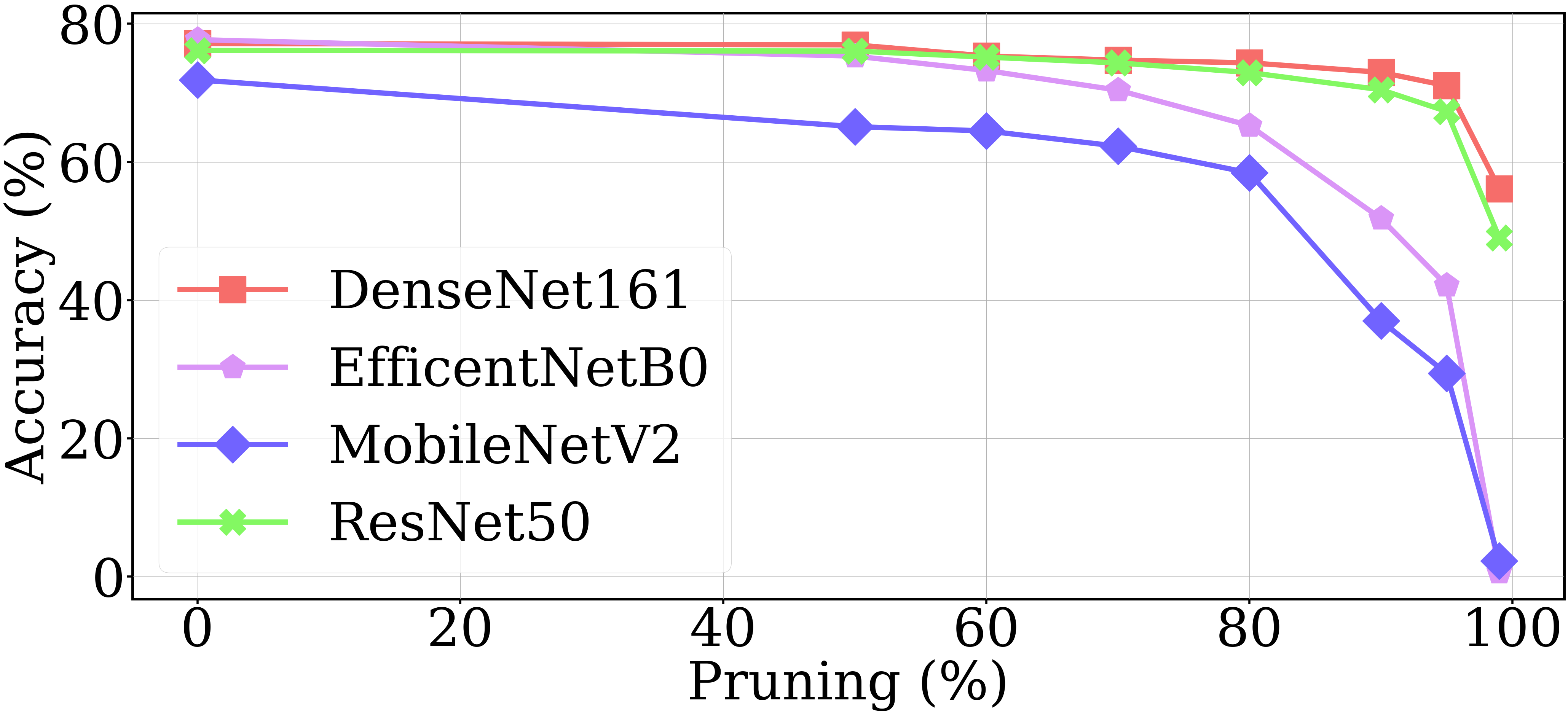}
  \caption{\footnotesize Weight pruning (ImageNet)}
  \label{fig:acc_unstructured_imagenet}
 \end{subfigure}
 \hspace{1em}%
\begin{subfigure}{0.31\textwidth}
  \centering
  \includegraphics[width=\linewidth]{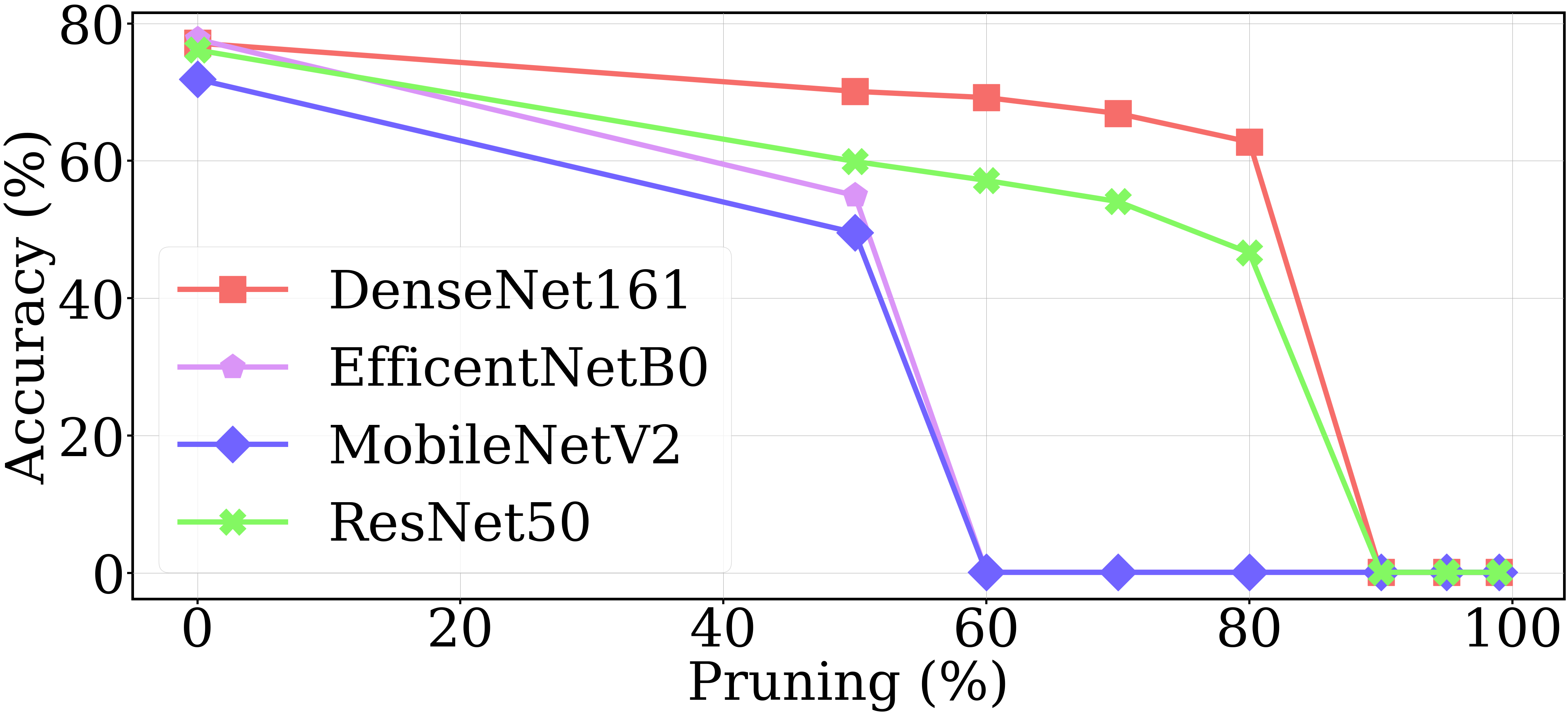}
  \caption{\footnotesize Channel pruning (ImageNet)}
  \label{fig:acc_structured_imagenet}
\end{subfigure}
 \hspace{1em}%
\begin{subfigure}{0.31\textwidth}
  \centering
\includegraphics[width=\linewidth]{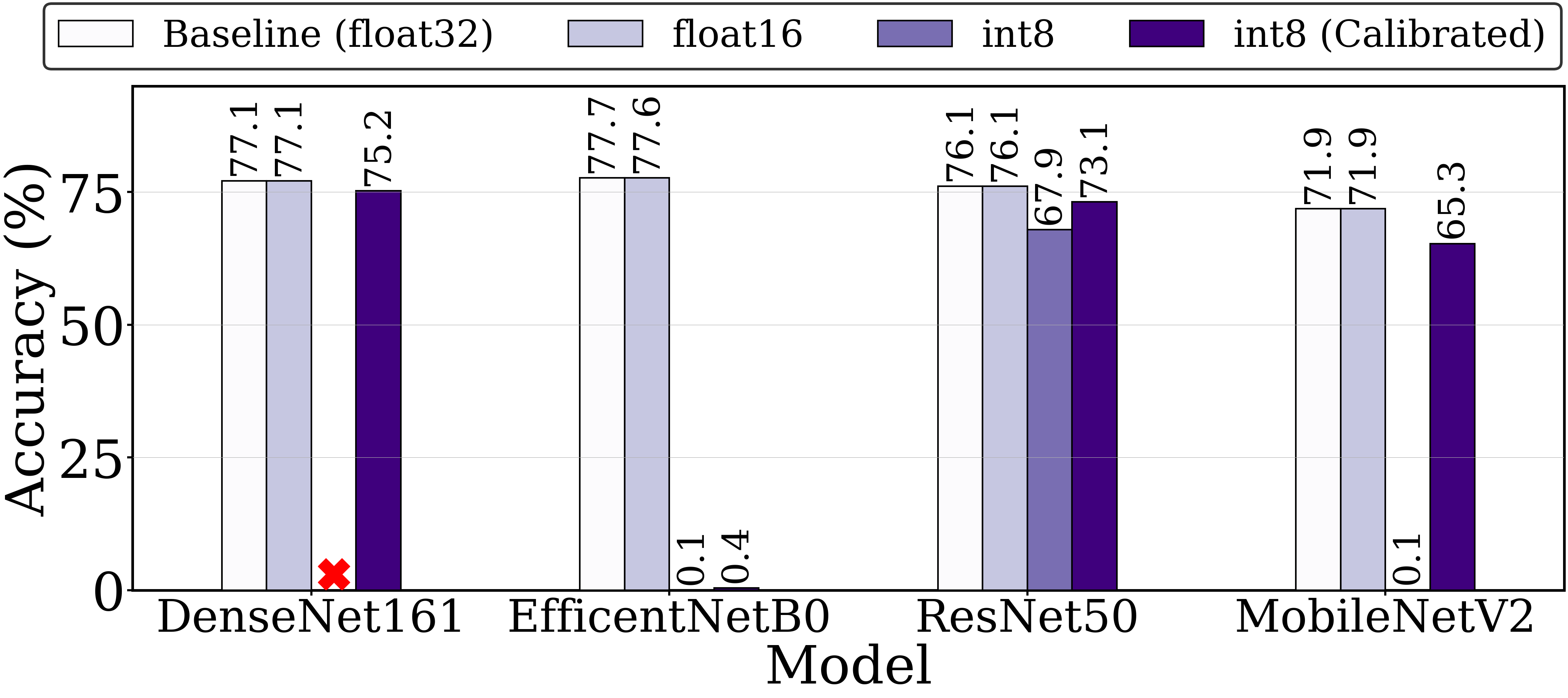}
  \caption{\footnotesize Data-type quantization (ImageNet)}
  \label{fig:acc_quant_imagenet}
\end{subfigure}
\caption{Accuracy and compression trade-offs for our CIFAR-10 (a-c) and ImageNet (d-f) models: (a/d) shows the accuracy of each model after iterative weight pruning of the convolutional layers, and fine-tuning the model; (b/e) shows a similar setup for channel pruning; and (c/f) shows \texttt{float16} and \texttt{int8} quantization accuracy, with \texttt{int8} accuracy shown before and after calibration.}
\label{fig:accuracies}
\end{figure*}

\begin{table}[t]
  \caption{CIFAR-10 models, including baseline accuracy (Top1), and our chosen compression ratios and corresponding accuracies.
    \label{tab:acc_cifar10}
  }
\centering
\footnotesize
\begin{tabular}{lccccccc}
\toprule
\multirow{2}{*}{Model} & \multirow{2}{*}{Params} & \multirow{2}{*}{MACs} & \multirow{2}{*}{Top1} & \multicolumn{4}{c}{Model Optimization Accuracy (\& Compression Ratio)}      \\ \cline{5-8}
                       &                         &                       &                       & Weight Pruning  & Channel Pruning  & \texttt{float16} & \texttt{int8} \\ \midrule
MobileNetV2            & 2.3M                    & 98M                   & 93.2\%                & 92.7\% (95\%)   & 89.3\% (50\%)    & 93.2\% (50\%) & 91.7\% (75\%)     \\
ResNet18               & 11.2M                   & 557M                  & 94.3\%                & 94.7\% (95\%)   & 89.0\% (80\%)    & 94.3\% (50\%) & 93.2\% (75\%)   \\
VGG-16                 & 14.7M                   & 314M                  & 92.9\%                & 93.1\% (95\%)   & 85.7\% (80\%)    & 92.9\% (50\%) & 92.0\% (75\%) \\
MobileNetV1            & 3.2M                    & 48M                   & 91.3\%                & 90.9\% (95\%)   & 86.9\% (50\%)    & 91.3\% (50\%) & 89.6\% (75\%)   \\

\bottomrule
\end{tabular}
\end{table}

\subsection{CIFAR-10}
\label{subsec:eval:cifar10}

\subsubsection{Accuracy}

The models' accuracy with varying levels of compression can be seen in the first row of Figure~\ref{fig:accuracies}.
For the four models, Table~\ref{tab:acc_cifar10} shows the baseline (dense) top-1 accuracy on CIFAR-10.
We observe that for weight pruning (Figure~\ref{fig:acc_unstructured_cifar10}), the accuracy is maintained for all models until 95\% pruning, at which point all models see a drop in accuracy at 99\%.
However, the drop in accuracy for MobileNetV1 and V2 is higher, likely because they have fewer parameters.

We also observe this trend in channel pruning (Figure~\ref{fig:acc_structured_cifar10}), where VGG-16 and ResNet18 maintain their accuracy for longer compared to the MobileNets.
However, for all models the drop in accuracy is earlier compared to weight pruning, with the elbow appearing at 50\% pruning for the MobileNet models, and 80\% for VGG-16 and ResNet18.
We also observe that after the elbow the drop is higher, to around 10\% accuracy or equivalent to random guessing.

For data-type quantization in Figure~\ref{fig:acc_quant_cifar10}, we observe almost no change in accuracy for \texttt{float16} across the four models.
The output is not bit-wise identical, however at most this represents a 0.03\% difference in top-1 accuracy.
For uncalibrated \texttt{int8} quantization, all models see a drop in accuracy, with MobileNets V1 and V2 seeing the highest drops, to 10.0\% and 16.4\% respectively.
However, with calibration all models recover significantly, with MobileNetV1 losing the most accuracy at 1.7\%.
Table~\ref{tab:acc_cifar10} shows the elbow points of accuracy we take for the inference experiments.

\subsubsection{Inference -- CPU (untuned)}
\label{subsubsec:eval_cpu_inf_untuned}

\begin{figure*}[t]
  \centering
  \begin{subfigure}{0.95\textwidth}
    \centering  
    \hspace*{0.2cm}
    \includegraphics[width=\textwidth]{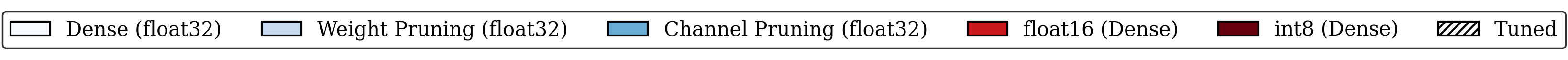}
  \end{subfigure}%
  \\
\centering
  \begin{subfigure}{0.24\textwidth}
    \centering     \includegraphics[width=\textwidth]{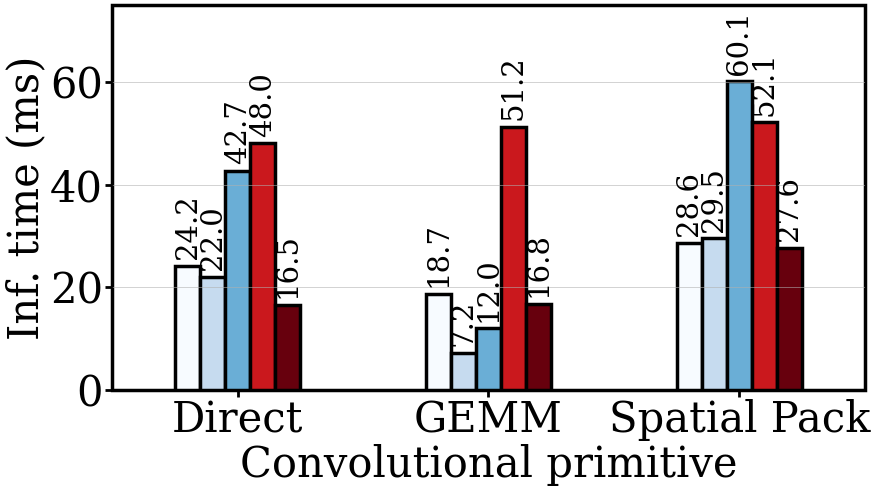}
    \caption{\footnotesize MobileNetV1 on HiKey} \label{fig:baseline_hikey_mobilenetv1}
  \end{subfigure}%
  \hspace*{\fill}
  \begin{subfigure}{0.24\textwidth}
    \centering     \includegraphics[width=\textwidth]{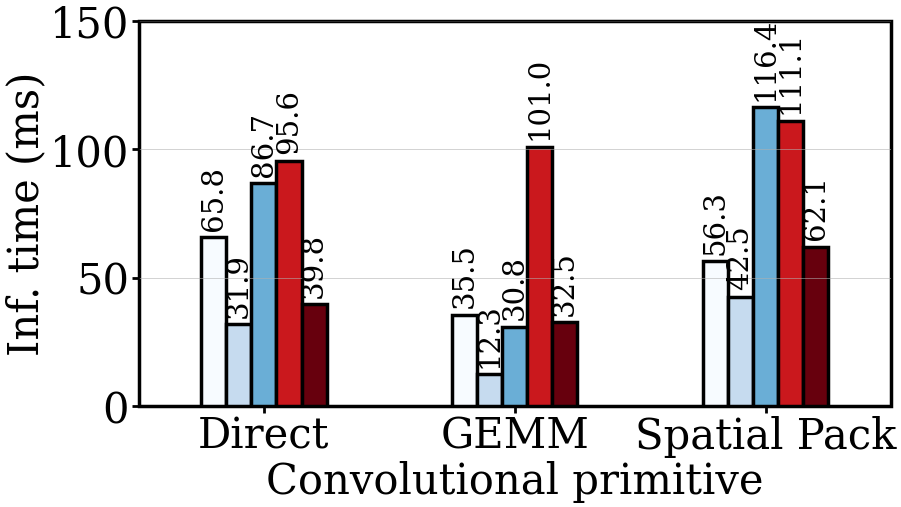}
    \caption{\footnotesize MobileNetV2 on HiKey} \label{fig:baseline_hikey_mobilenetv2}
  \end{subfigure}%
  \hspace*{\fill}
  \begin{subfigure}{0.24\textwidth}
    \centering     \includegraphics[width=\textwidth]{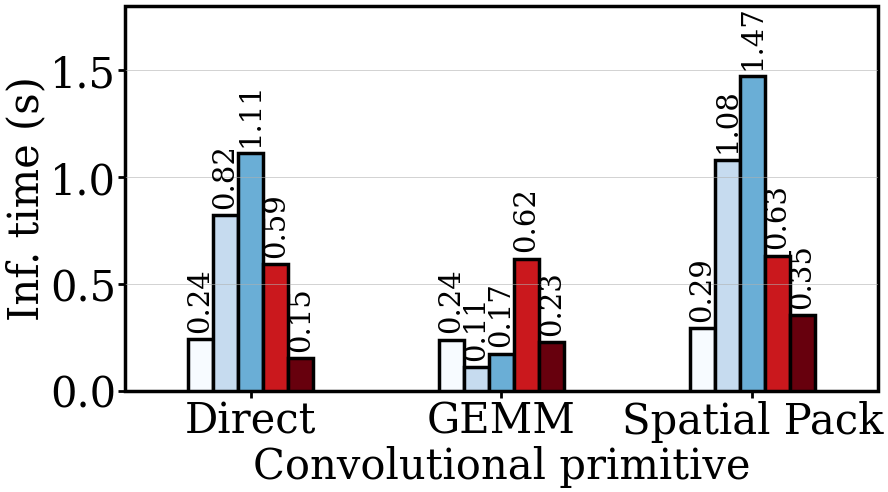}
    \caption{\footnotesize ResNet18 on HiKey} \label{fig:baseline_hikey_resnet18}
  \end{subfigure}%
  \hspace*{\fill}
\begin{subfigure}{0.24\textwidth}
    \centering     \includegraphics[width=\textwidth]{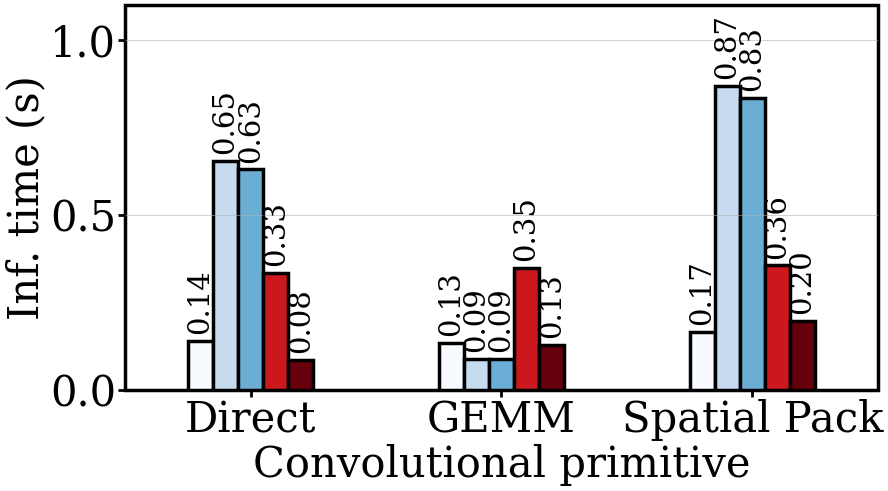}
    \caption{\footnotesize VGG-16 on HiKey} \label{fig:baseline_hikey_vgg}
  \end{subfigure}%
  \\
    \begin{subfigure}{0.24\textwidth}
    \centering     \includegraphics[width=\textwidth]{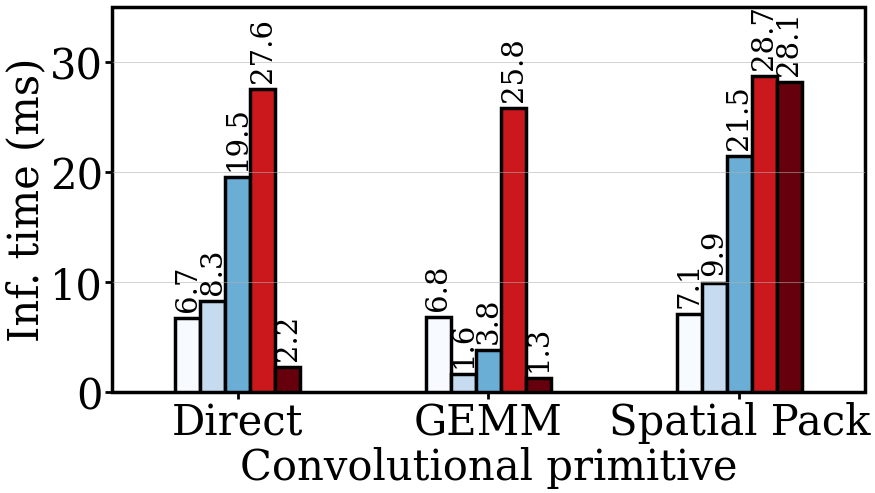}
    \caption{\footnotesize MobileNetV1 on Intel i7} \label{fig:baseline_i7_mobilenetv1}
  \end{subfigure}
    \hspace*{\fill}   %
  \begin{subfigure}{0.24\textwidth}
    \centering     \includegraphics[width=\textwidth]{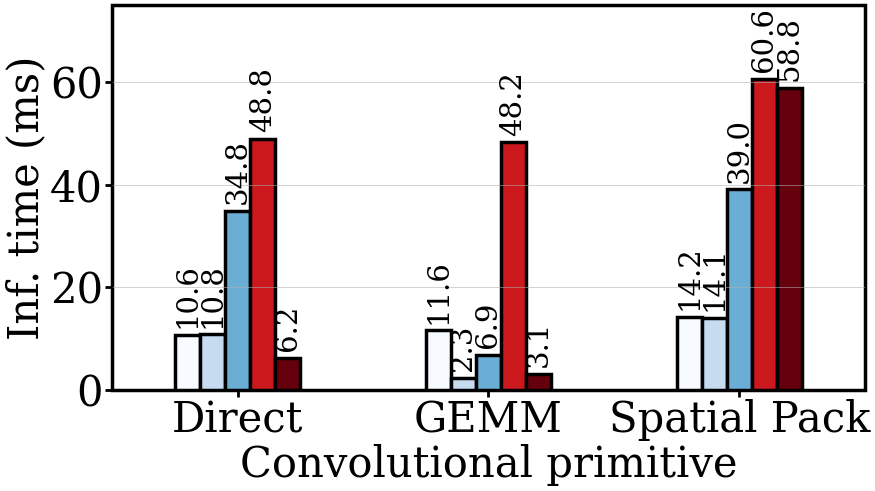}
    \caption{\footnotesize MobileNetV2 on Intel i7} \label{fig:baseline_i7_mobilenetv2}
  \end{subfigure}
    \hspace*{\fill}   %
   \begin{subfigure}{0.24\textwidth}
    \centering     \includegraphics[width=\textwidth]{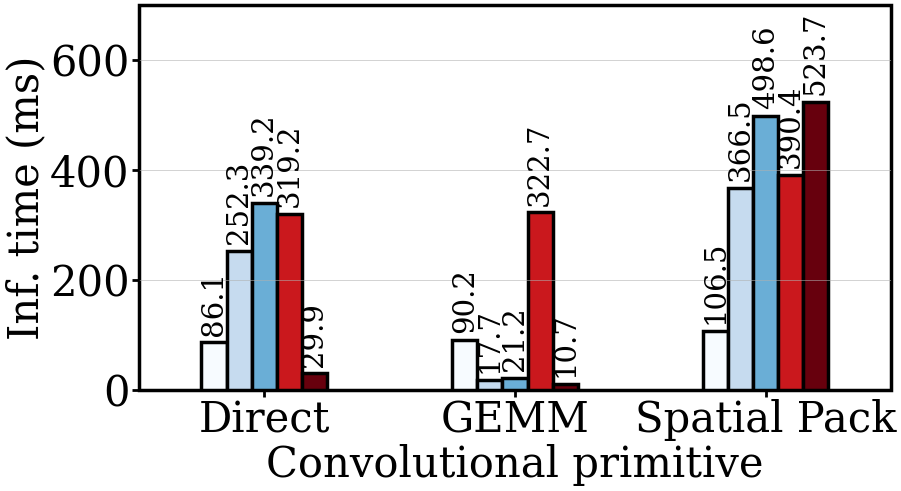}
    \caption{\footnotesize ResNet18 on Intel i7} \label{fig:baseline_i7_resnet18}
  \end{subfigure}%
    \hspace*{\fill}   %
    \begin{subfigure}{0.24\textwidth}
    \centering     \includegraphics[width=\textwidth]{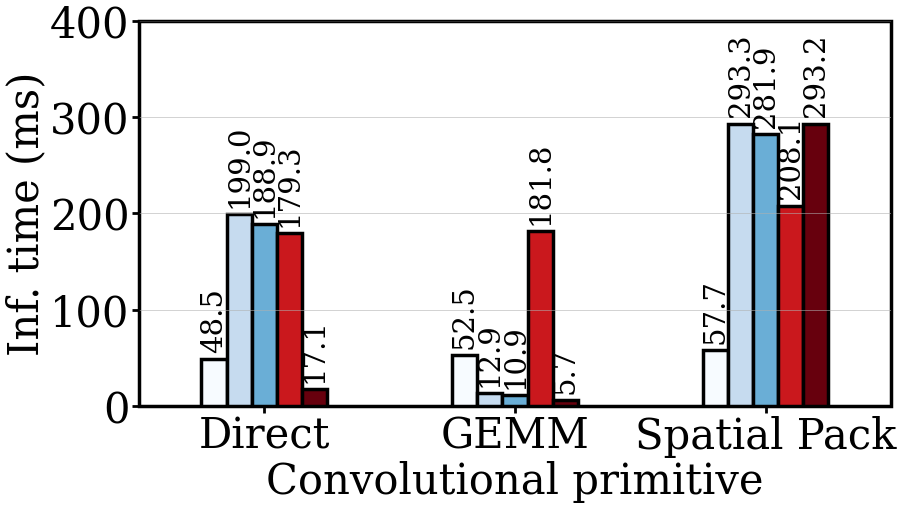}
    \caption{\footnotesize VGG-16 on Intel i7} \label{fig:baseline_i7_vgg}
  \end{subfigure}%
  \\
  \begin{subfigure}{0.24\textwidth}
    \centering     \includegraphics[width=\textwidth]{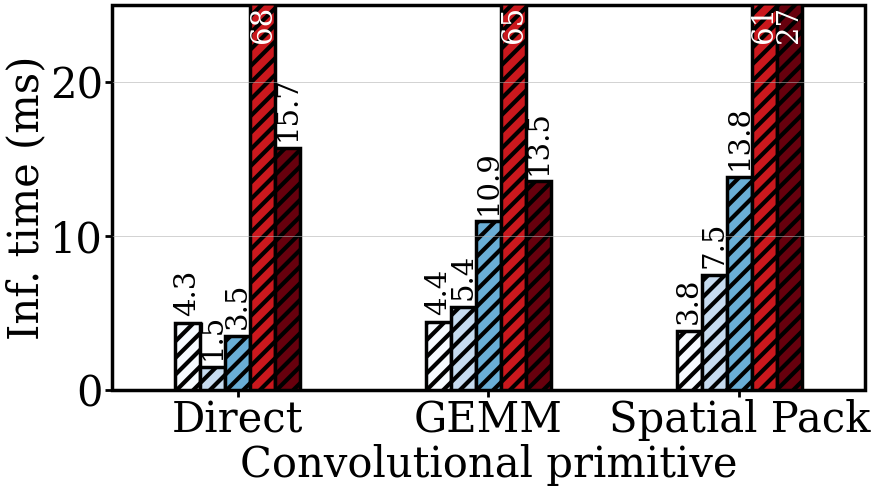}
    \caption{\footnotesize MobileNetV1 on HiKey} \label{fig:cpu_tuned_hikey_mobilenetv1}
  \end{subfigure}%
  \hspace*{\fill}   %
   \begin{subfigure}{0.24\textwidth}
    \centering     \includegraphics[width=\textwidth]{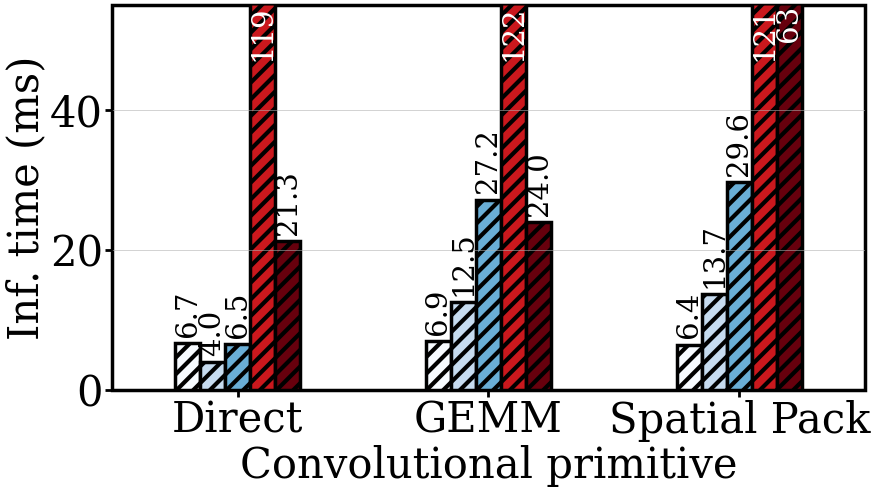}
    \caption{\footnotesize MobileNetV2 on HiKey} \label{fig:cpu_tuned_hikey_mobilenetv2}
  \end{subfigure}%
    \hspace*{\fill}   %
     \begin{subfigure}{0.24\textwidth}
    \centering     \includegraphics[width=\textwidth]{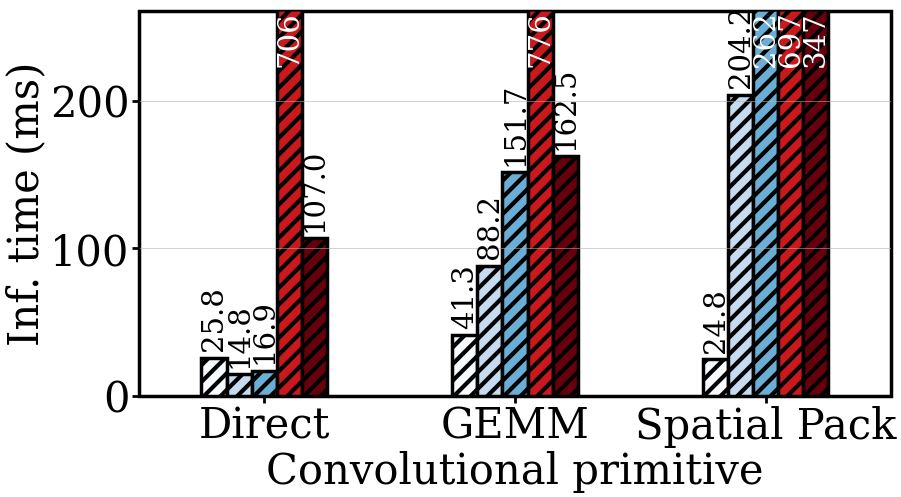}
    \caption{\footnotesize ResNet18 on HiKey} \label{fig:cpu_tuned_hikey_resnet18}
  \end{subfigure}%
    \hspace*{\fill}   %
  \begin{subfigure}{0.24\textwidth}
    \centering     \includegraphics[width=\textwidth]{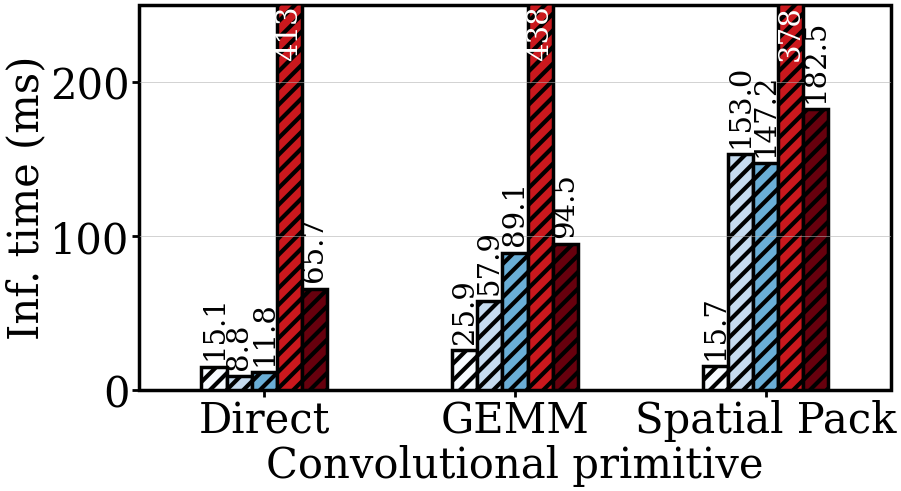}
    \caption{\footnotesize VGG-16 on HiKey} \label{fig:cpu_tuned_hikey_vgg}
  \end{subfigure}%
  \\
  \begin{subfigure}{0.24\textwidth}
    \centering     \includegraphics[width=\textwidth]{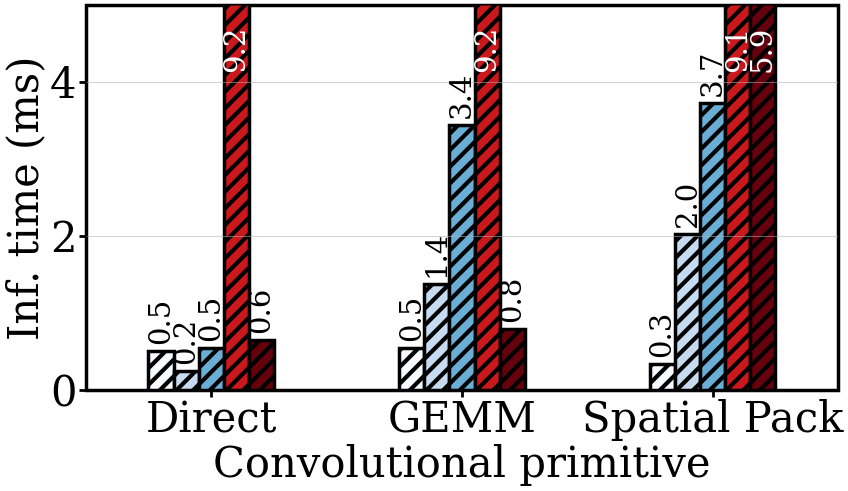}
    \caption{\footnotesize MobileNetV1 on Intel i7} \label{fig:cpu_tuned_i7_mobilenetv1}
  \end{subfigure}
     \hspace*{\fill}   %
 \begin{subfigure}{0.24\textwidth}
    \centering     \includegraphics[width=\textwidth]{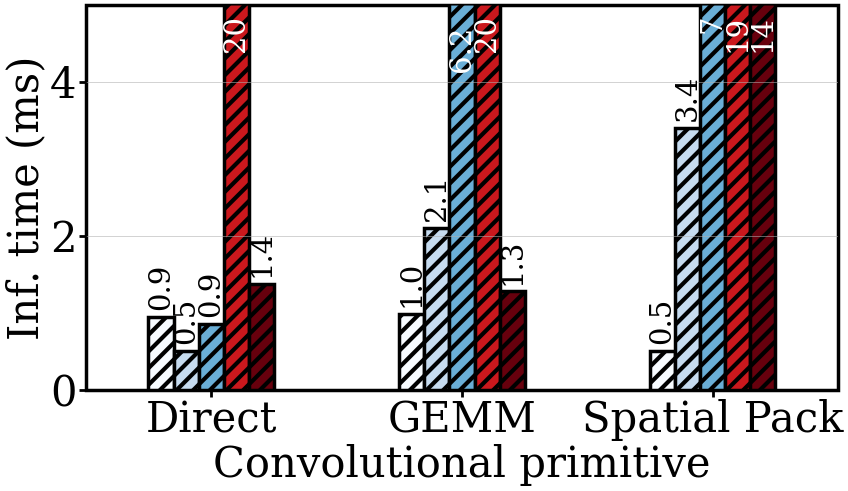}
    \caption{\footnotesize MobileNetV2 on Intel i7} \label{fig:cpu_tuned_i7_mobilenetv2}
  \end{subfigure}
  \hspace*{\fill}   %
    \begin{subfigure}{0.24\textwidth}
    \centering     \includegraphics[width=\textwidth]{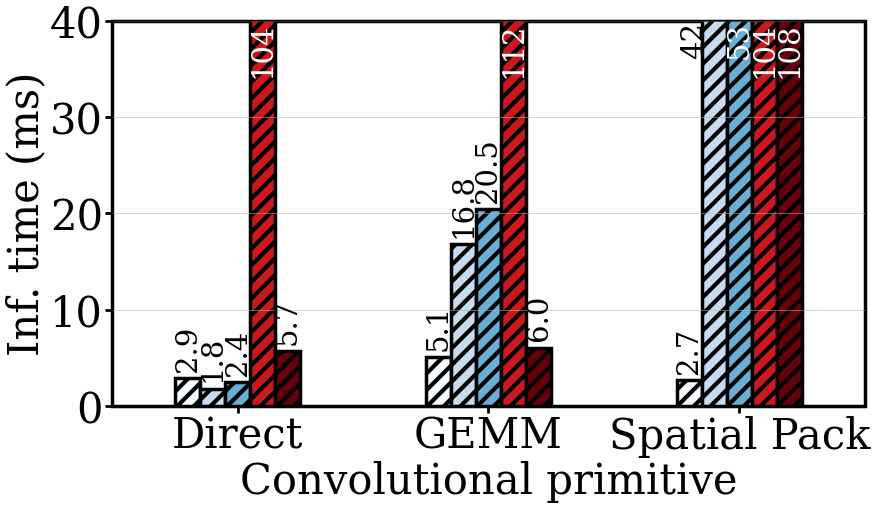}
    \caption{\footnotesize ResNet18 on Intel i7} \label{fig:cpu_tuned_i7_resnet18}
  \end{subfigure}%
  \hspace*{\fill}   %
    \begin{subfigure}{0.24\textwidth}
    \centering     \includegraphics[width=\textwidth]{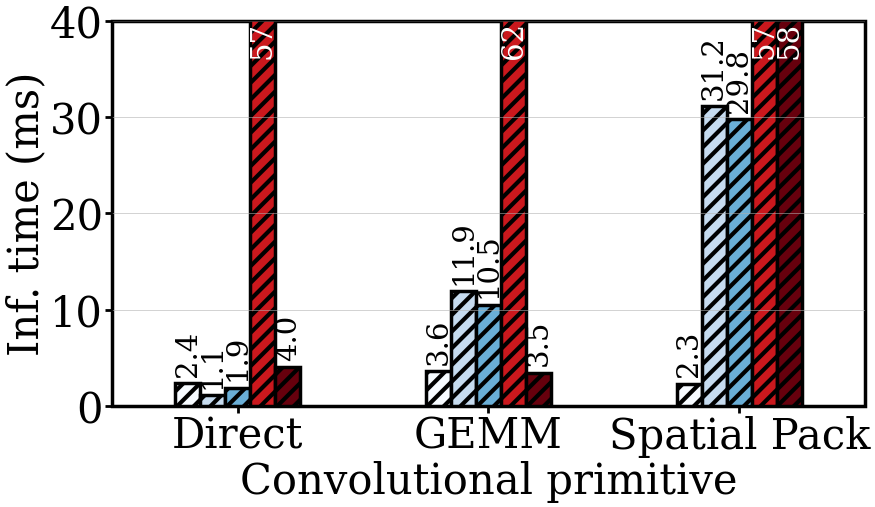}
    \caption{\footnotesize VGG-16 on Intel i7} \label{fig:cpu_tuned_i7_vgg}
  \end{subfigure}%
    \caption{Experiments comparing the compressed CIFAR-10 models chosen from obvious elbows of accuracy, with varying algorithmic primitives, benchmarked on the i7 and HiKey CPU platforms, with and without auto-scheduling.}
  	\label{fig:inference_cifar10_cpu}
\end{figure*}

The first two rows of Figure~\ref{fig:inference_cifar10_cpu} show the untuned performance of the CIFAR-10 models when running on the CPUs of the HiKey (\ref{fig:baseline_hikey_mobilenetv1}-\ref{fig:baseline_hikey_vgg}) and i7 (\ref{fig:baseline_i7_mobilenetv1}-\ref{fig:baseline_i7_vgg}) platforms, with varying compression strategies and convolutional primitives.
The overall trends, including the fastest combination of parameters under different settings, are shown in Table~\ref{tab:cifar10_results}.
For the untuned baseline, we observe that the i7 and HiKey differ in fastest algorithm, with \emph{direct} and \emph{GEMM} for all models respectively.
However, in most cases across devices \emph{GEMM} algorithms are the fastest, with one exception for VGG-16 on the HiKey.
In terms of compression techniques, \texttt{int8} has the fastest overall time in three out of four cases on the i7, whereas weight pruning is the fastest on the Hikey in three out of four cases.

If we take the \emph{best} baseline time for each model, we can compute an expected speedup given the compression ratio of each model optimization technique.
For example, on the HiKey for MobileNetV1, with a pruning rate of 95\% we could expect an ideal speedup of 20$\times$.
However, in this case we only achieve a speedup of 2.6$\times$ for the best weight pruning algorithm (\emph{GEMM}), i.e., 13.0\% of the expected speedup.
On average, for weight pruning we achieve 11.5\% and 21.8\% of the expected speedup on the HiKey and i7 respectively.

For channel pruning, the elbow points for the models (see Table~\ref{tab:acc_cifar10}) show that channel pruned models are less compressed than the weight pruning models, which means that we would the latter to always be slower than the former.
However, we observe several cases where a channel pruning model is \emph{faster}, namely for all VGG-16 variants, except for \emph{GEMM} on the HiKey, which is slightly slower.
On average, for channel pruning, we achieve 77.9\% and 83.9\% of the expected speedup on the HiKey and i7 respectively.

For \texttt{float16}, we observe a slowdown when compared to the baseline (\texttt{float32}) in every case.
For \texttt{int8}, we generally see a speedup relative to the baseline, with some notable exceptions using \emph{spatial pack}.
In some cases, \texttt{int8} gives the best time overall, as seen in the last column of Table~\ref{tab:cifar10_results}.
On average, we achieve 33.6\% and 157.2\% of the expected speedup on the HiKey and i7 respectively.

\begin{table}[]
\footnotesize
\caption{Analysis of CIFAR-10 CPU results for varying combinations of parameters, summarizing Figure~\ref{fig:inference_cifar10_cpu}.
The best inference times are shown in \textbf{bold}.
Shortened names are used for brevity: WP (weight pruning), CP (channel pruning), i8 (\texttt{int8}), f16 (\texttt{float16}).
\label{tab:cifar10_results}}
\begin{tabular}{cl|ccccc|ccc|c}
\toprule
\multirow{2}{*}{\textbf{Platform}} & \multirow{2}{*}{\textbf{Model}} & \multicolumn{5}{|c|}{\textbf{Fastest algorithm}}             & \multicolumn{3}{|c|}{\textbf{Fastest compression technique}} & \multirow{2}{*}{\shortstack{\textbf{Overall}\\\textbf{fastest}}} \\
                                   &                                 & Dense & WP            & CP   & i8              & f16    & GEMM              & Direct            & Spatial (Pack)       &                                   \\
                                   \hline
\multirow{4}{*}{\shortstack{HiKey \\ (untuned)}}             & MobileNetV2                     & GEMM     & \textbf{GEMM} & GEMM & GEMM            & Direct & \textbf{WP}       & WP                & WP                 & WP+GEMM                           \\
                                   & ResNet18                        & GEMM     & \textbf{GEMM} & GEMM & Direct          & Direct & \textbf{WP}       & i8                & Dense           & WP+GEMM                           \\
                                   & VGG-16                          & GEMM     & GEMM          & GEMM & \textbf{Direct} & Direct & WP                & \textbf{i8}       & Dense           & i8+Direct                         \\
                                   & MobileNetV1                     & GEMM     & \textbf{GEMM} & GEMM & Direct          & Direct & \textbf{WP}       & i8                & i8                 & WP+GEMM
                                   \\ \hline
\multirow{4}{*}{\shortstack{i7 \\ (untuned)}}                & MobileNetV2                     & Direct   & \textbf{GEMM} & GEMM & GEMM            & GEMM   & \textbf{WP}       & i8                & WP                 & WP+GEMM                           \\
                                   & ResNet18                        & Direct   & GEMM          & GEMM & \textbf{GEMM}   & Direct & \textbf{i8}       & i8                & Dense           & i8+GEMM                           \\
                                   & VGG-16                          & Direct   & GEMM          & GEMM & \textbf{GEMM}   & Direct & \textbf{i8}       & i8                & Dense           & i8+GEMM                           \\
                                   & MobileNetV1                     & Direct   & GEMM          & GEMM & \textbf{GEMM}   & GEMM   & \textbf{i8}       & i8                & Dense           & i8+GEMM
                                   \\
  \hline
\multirow{4}{*}{\shortstack{HiKey \\ (tuned)}}             & MobileNetV2                     & Spatial          & \textbf{Direct} & Direct & Direct & Direct  & Dense         & \textbf{WP}        & Dense                 & WP+Direct                         \\
                                   & ResNet18                        & Spatial          & \textbf{Direct} & Direct & Direct & Spatial & Dense         & \textbf{WP}        & Dense                 & WP+Direct                         \\
                                   & VGG-16                          & Direct           & \textbf{Direct} & Direct & Direct & Spatial & Dense         & \textbf{WP}        & Dense                 & WP+Direct                         \\
                                   & MobileNetV1                     & Spatial          & \textbf{Direct} & Direct & GEMM   & Spatial & Dense         & \textbf{WP}        & Dense                 & WP+Direct
  \\ \hline
\multirow{4}{*}{\shortstack{i7 \\ (tuned)}}                & MobileNetV2                & \textbf{Spatial} & Direct          & Direct & GEMM   & Spatial & Dense         & WP                 & \textbf{Dense}        & Dense+Spatial                     \\
                                   & ResNet18                        & Spatial          & \textbf{Direct} & Direct & Direct & Spatial & Dense         & \textbf{WP}        & Dense                 & WP+Direct                         \\
                                   & VGG-16                          & Spatial          & \textbf{Direct} & Direct & GEMM   & Direct  & i8            & \textbf{WP}        & Dense                 & WP+Direct                         \\
                                   & MobileNetV1                     & Spatial          & \textbf{Direct} & Direct & Direct & Direct  & Dense         & \textbf{WP}        & Dense                 & WP+Direct                         \\
                                   \bottomrule
\end{tabular}
\end{table}

\subsubsection{Inference -- CPU (tuned)}
\label{subsubsec:eval_cpu_inf_ansor}

The last two rows of Figure~\ref{fig:inference_cifar10_cpu} show the tuned performance of the CIFAR-10 models when running on the HiKey (\ref{fig:cpu_tuned_hikey_mobilenetv1}-\ref{fig:cpu_tuned_hikey_vgg}) and i7 (\ref{fig:cpu_tuned_i7_mobilenetv1}-\ref{fig:cpu_tuned_i7_vgg}) CPUs, with overall trends shown in Table~\ref{tab:cifar10_results}.
Comparing to the untuned results in Section~\ref{subsubsec:eval_cpu_inf_untuned}, we observe that tuning has created some significant differences in the relative performance of the experiments, beyond reducing the inference time significantly.
For example, \emph{spatial pack} is now predominantly the best algorithm for dense models on both CPUs.
For weight pruning, \emph{direct} is the best algorithm and this combination is predominantly the fastest inference time across models and CPUs.
On average, we achieve 9.5\% and 6.9\% of the expected speedup on the HiKey and i7 respectively, less than untuned but faster in absolute terms.
For channel pruning, sparse \emph{direct} is also the best algorithm, and we still observe that channel pruning is faster than weight pruning in most cases of VGG-16.
On average, we achieve 39.8\% and 26.6\% of the expected speedup on the HiKey and i7 respectively.
Also, \texttt{float16} still gives a consistent slowdown, and these differences are exacerbated when compared to the untuned results.
The relative speedups of \texttt{int8} on the i7 have disappeared with tuning, with an average slowdown of 2.0$\times$.

\subsubsection{Inference -- GPU (untuned)}

\begin{figure*}[t]
  \centering
  \begin{subfigure}{0.95\textwidth}
    \centering     \hspace*{0.2cm}    \includegraphics[width=\textwidth]{fig_neo/inference/legend.png}
  \end{subfigure}%
  \\
\centering
  \begin{subfigure}{0.24\textwidth}
    \includegraphics[width=\textwidth]{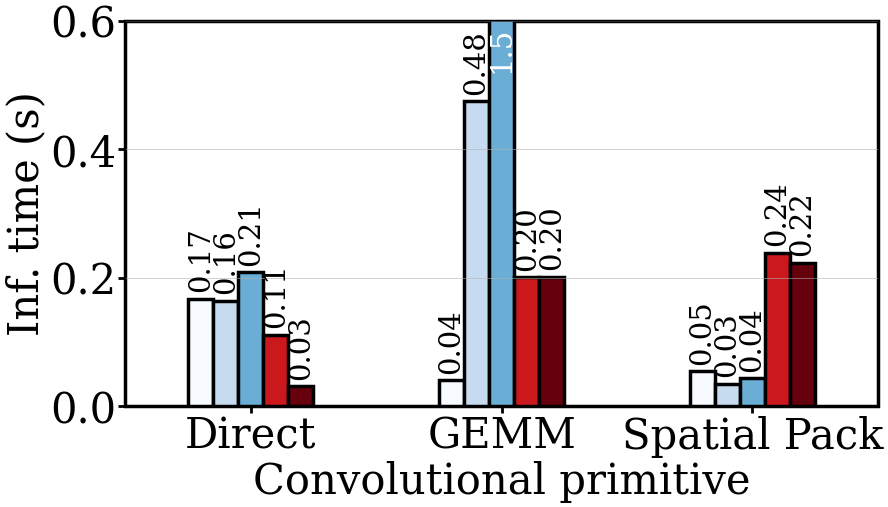}
    \caption{\footnotesize MobileNetV1 on HiKey} \label{fig:gpu_untuned_hikey_mobilenetv1}
  \end{subfigure}%
  \hspace*{\fill}
  \begin{subfigure}{0.24\textwidth}
    \includegraphics[width=\textwidth]{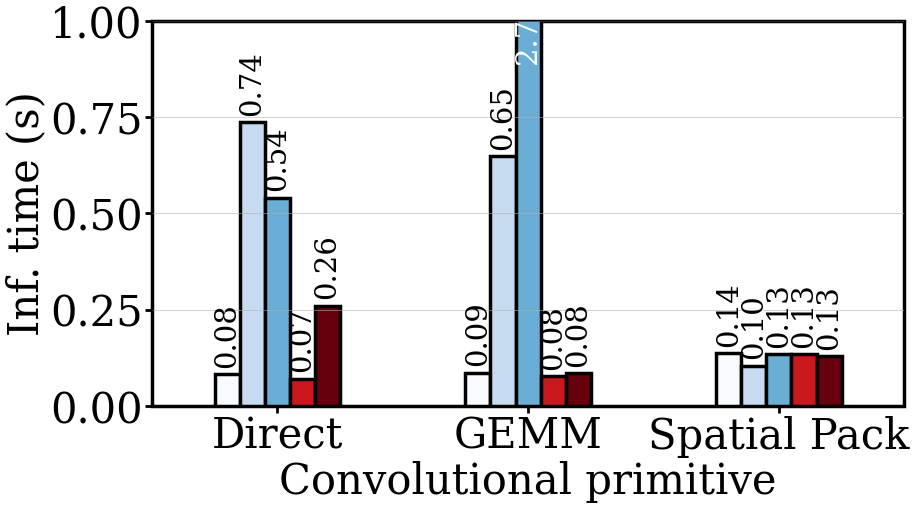}
    \caption{\footnotesize MobileNetV2 on HiKey} \label{fig:gpu_untuned_hikey_mobilenetv2}
  \end{subfigure}%
  \hspace*{\fill}
  \begin{subfigure}{0.24\textwidth}
    \includegraphics[width=\textwidth]{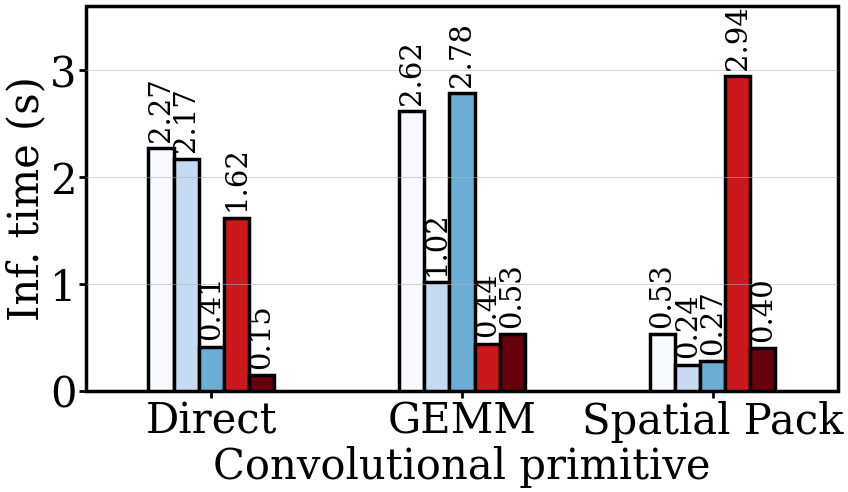}
    \caption{\footnotesize ResNet18 on HiKey} \label{fig:gpu_untuned_hikey_resnet18}
  \end{subfigure}%
  \hspace*{\fill}
  \begin{subfigure}{0.24\textwidth}
    \includegraphics[width=\textwidth]{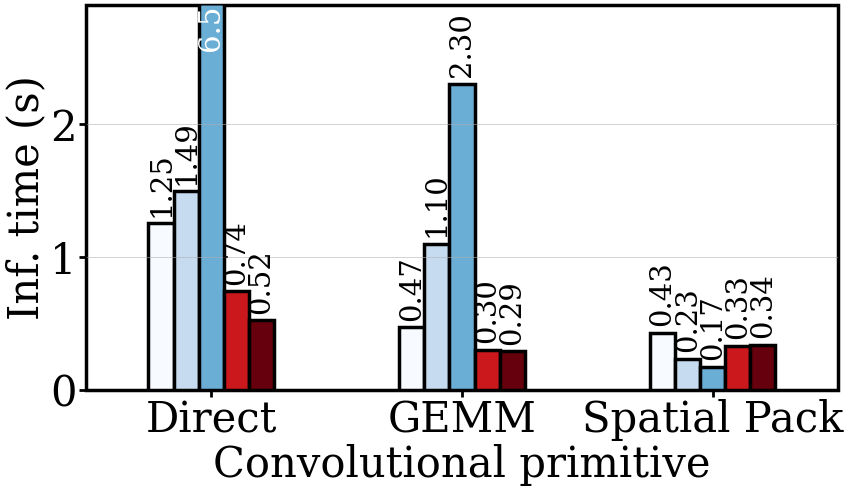}
    \caption{\footnotesize VGG-16 on HiKey} \label{fig:gpu_untuned_hikey_vgg}
  \end{subfigure}
  \\
   \begin{subfigure}{0.24\textwidth}
    \includegraphics[width=\textwidth]{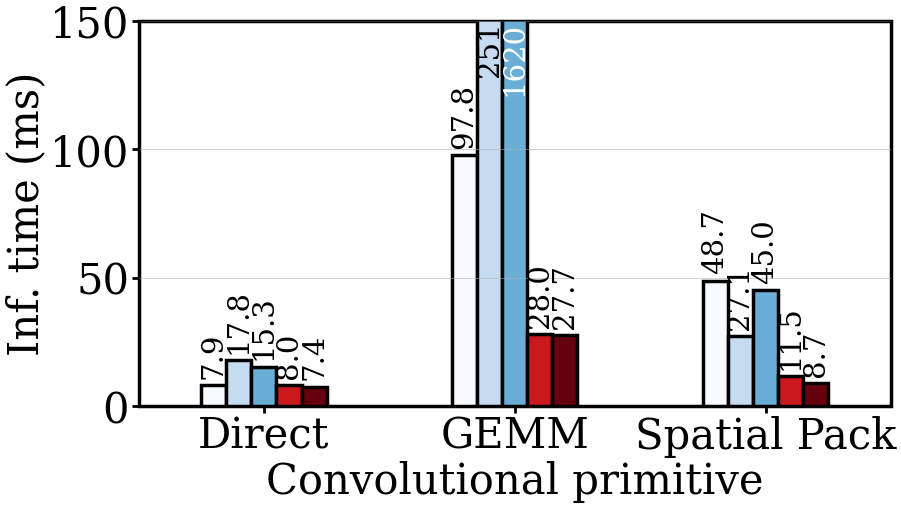}
    \caption{\footnotesize MobileNetV1 on Xavier} \label{fig:gpu_untuned_xavier_mobilenetv1}
  \end{subfigure}
    \hspace*{\fill}
  \begin{subfigure}{0.24\textwidth}
    \includegraphics[width=\textwidth]{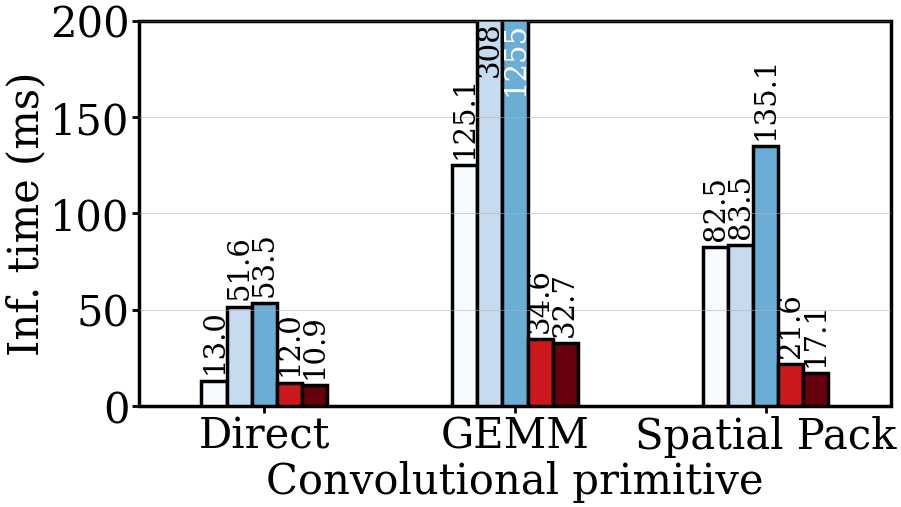}
    \caption{\footnotesize MobileNetV2 on Xavier} \label{fig:gpu_untuned_xavier_mobilenetv2}
  \end{subfigure}
    \hspace*{\fill}
  \begin{subfigure}{0.24\textwidth}
    \includegraphics[width=\textwidth]{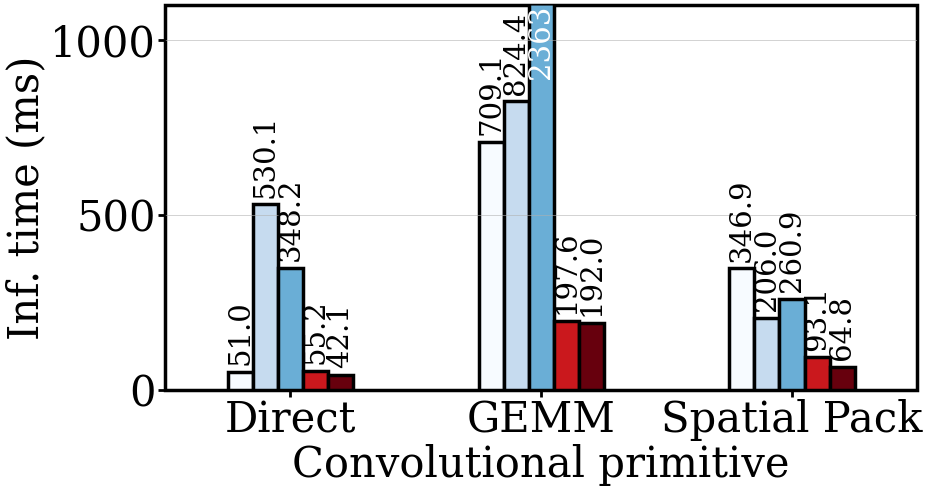}
    \caption{\footnotesize ResNet18 on Xavier} \label{fig:gpu_untuned_xavier_resnet18}
  \end{subfigure}%
    \hspace*{\fill}
    \begin{subfigure}{0.24\textwidth}
    \includegraphics[width=\textwidth]{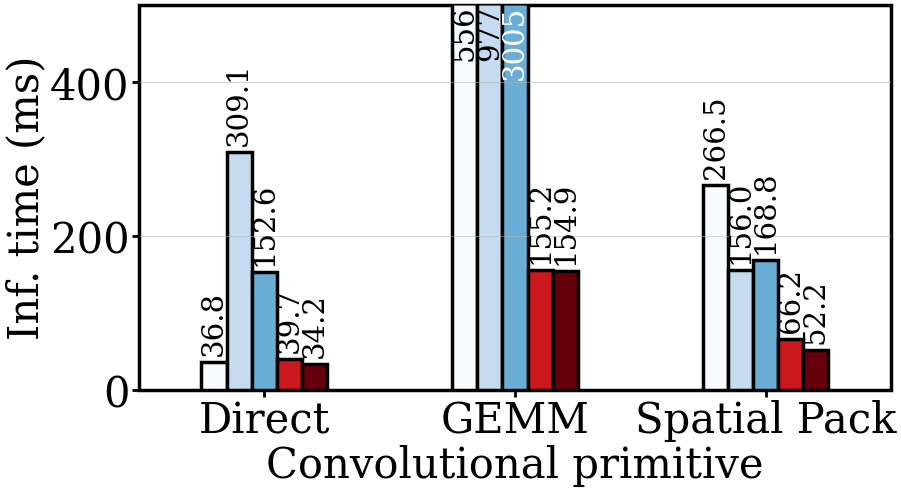}
    \caption{\footnotesize VGG-16 on Xavier} \label{fig:gpu_untuned_xavier_vgg}
  \end{subfigure}%
\\
  \begin{subfigure}{0.24\textwidth}
    \includegraphics[width=\textwidth]{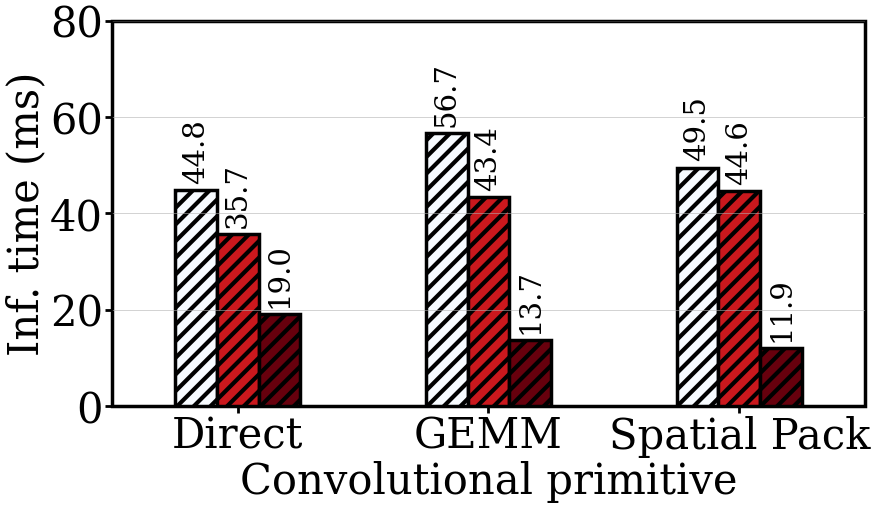}
    \caption{\footnotesize MobileNetV1 on HiKey} \label{fig:gpu_tuned_hikey_mobilenetv1}
  \end{subfigure}%
  \hspace*{\fill}
  \begin{subfigure}{0.24\textwidth}
    \includegraphics[width=\textwidth]{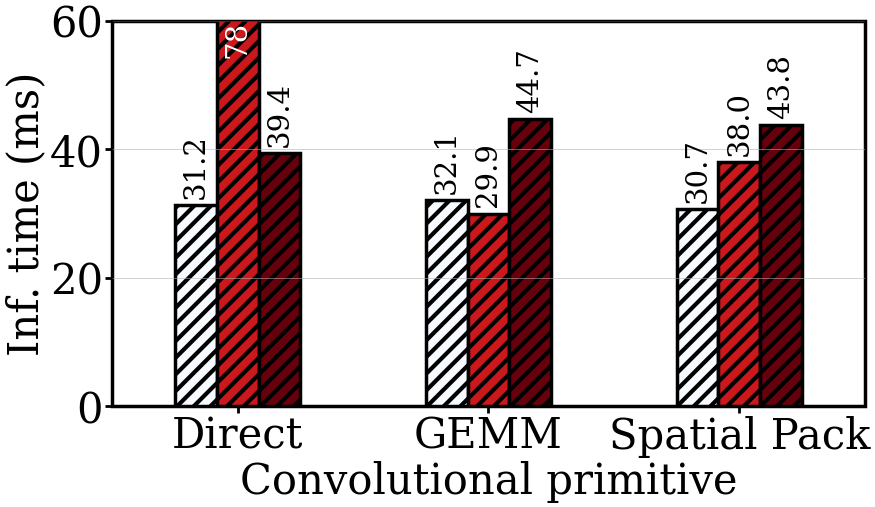}
    \caption{\footnotesize MobileNetV2 on HiKey} \label{fig:gpu_tuned_hikey_mobilenetv2}
  \end{subfigure}%
  \hspace*{\fill}
  \begin{subfigure}{0.24\textwidth}
    \includegraphics[width=\textwidth]{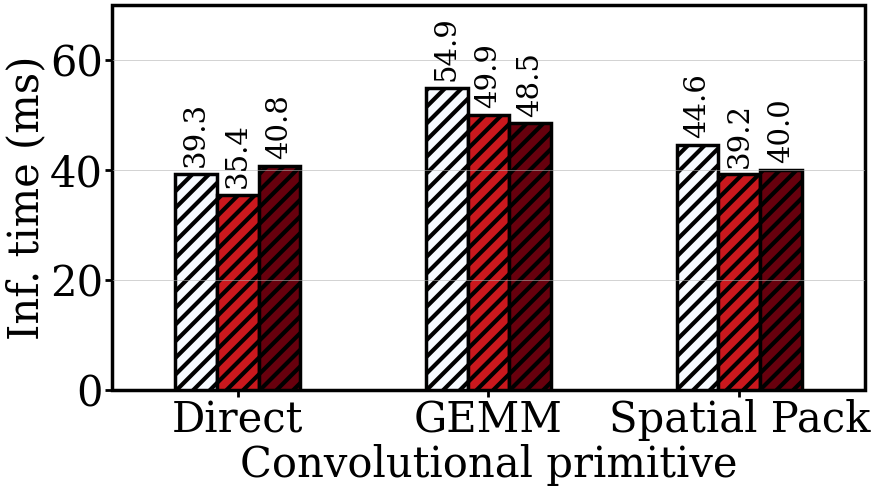}
    \caption{\footnotesize ResNet18 on HiKey} \label{fig:gpu_tuned_hikey_resnet18}
  \end{subfigure}%
  \hspace*{\fill}
  \begin{subfigure}{0.24\textwidth}
    \includegraphics[width=\textwidth]{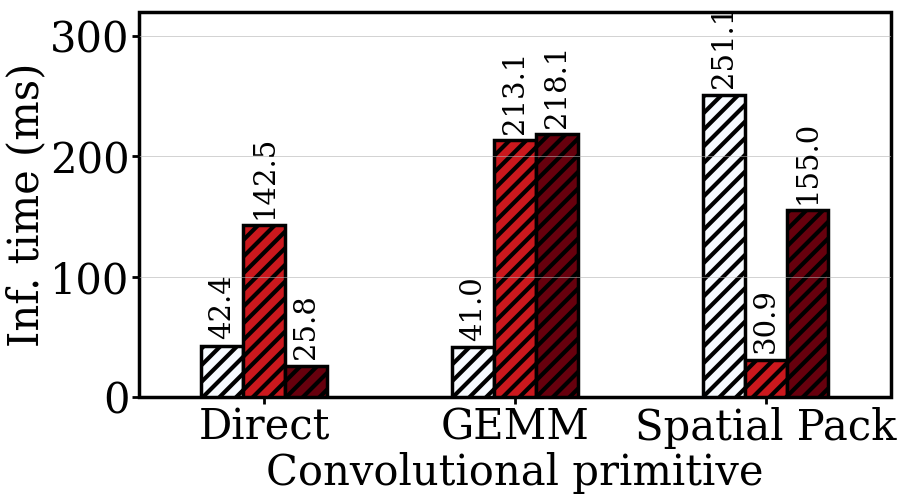}
    \caption{\footnotesize VGG-16 on HiKey} \label{fig:gpu_tuned_hikey_vgg}
  \end{subfigure}%
  \\
    \begin{subfigure}{0.24\textwidth}
    \includegraphics[width=\textwidth]{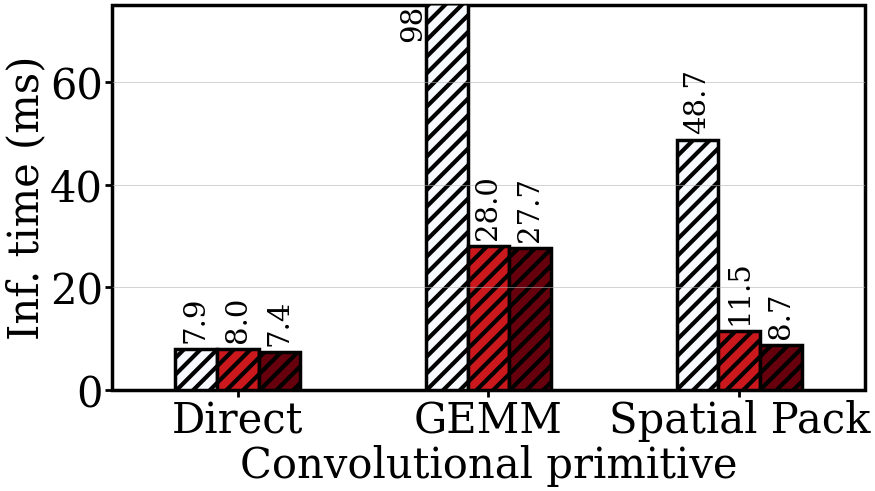}
    \caption{\footnotesize MobileNetV1 on Xavier} \label{fig:gpu_tuned_xavier_mobilenetv1}
  \end{subfigure}
   \hspace*{\fill}
    \begin{subfigure}{0.24\textwidth}
    \includegraphics[width=\textwidth]{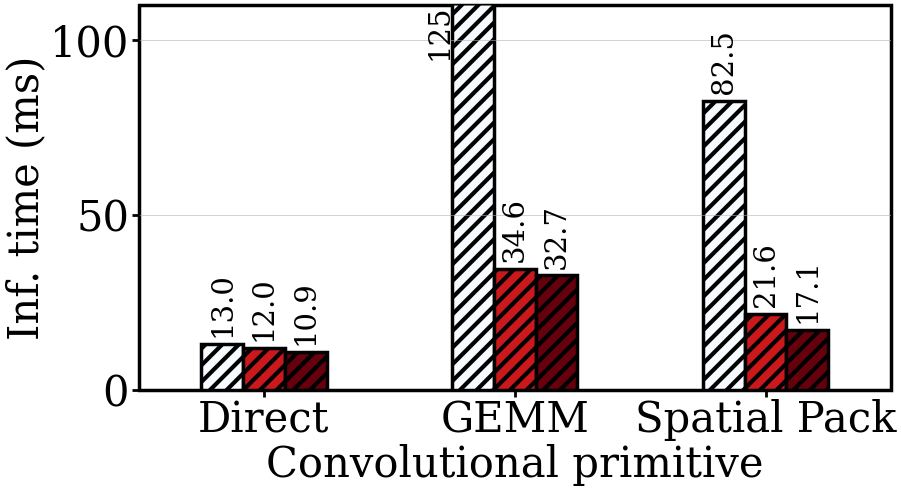}
    \caption{\footnotesize MobileNetV2 on Xavier} \label{fig:gpu_tuned_xavier_mobilenetv2}
  \end{subfigure}
    \hspace*{\fill}
    \begin{subfigure}{0.24\textwidth}
    \includegraphics[width=\textwidth]{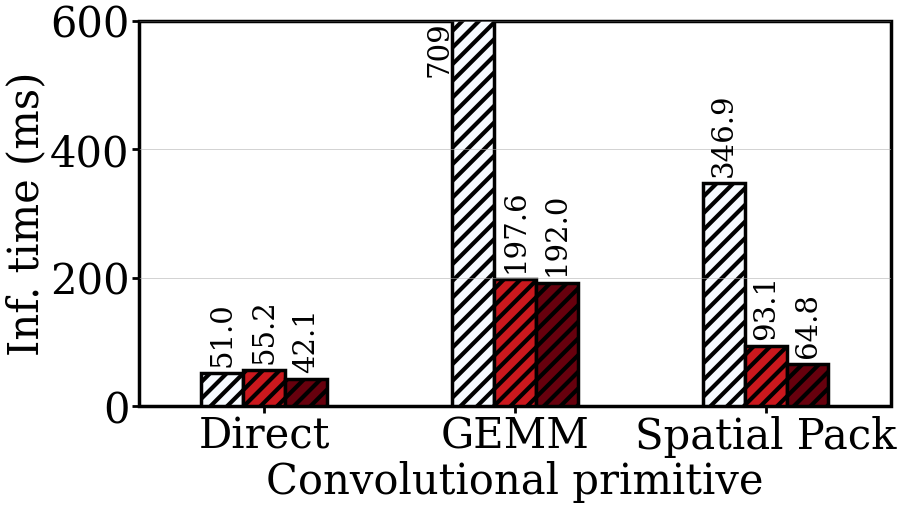}
    \caption{\footnotesize ResNet18 on Xavier} \label{fig:gpu_tuned_xavier_resnet18}
  \end{subfigure}%
     \hspace*{\fill}
    \begin{subfigure}{0.24\textwidth}
    \includegraphics[width=\textwidth]{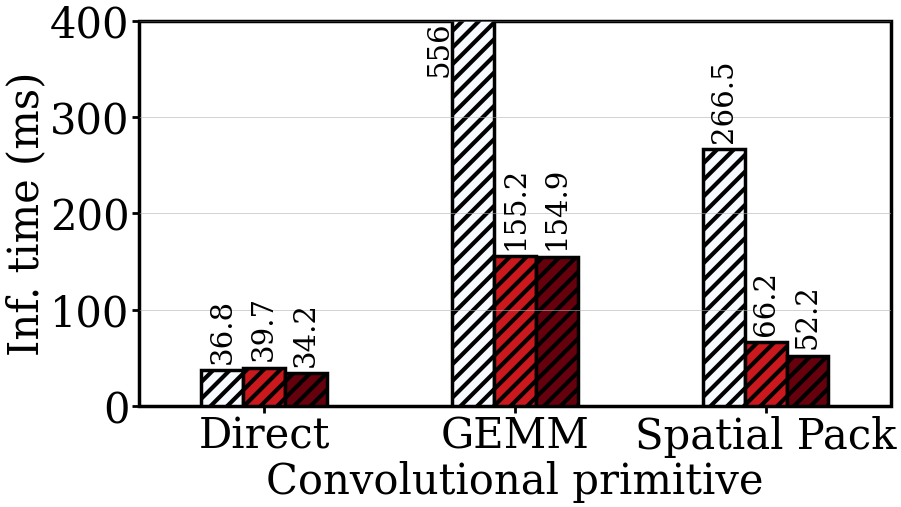}
    \caption{\footnotesize VGG-16 on Xavier} \label{fig:gpu_tuned_xavier_vgg}
  \end{subfigure}%
    \caption{Experiments comparing the compressed CIFAR-10 models chosen from obvious elbows of accuracy, with varying algorithmic primitives, benchmarked on the HiKey and Xavier GPU platforms, with and without auto-scheduling.}
  	\label{fig:inference_cifar10_gpu}
\end{figure*}

The first two rows of Figure~\ref{fig:inference_cifar10_gpu} show the untuned performance of the CIFAR-10 models when running on the GPUs of the HiKey (\ref{fig:gpu_untuned_hikey_mobilenetv1}-\ref{fig:gpu_untuned_hikey_vgg}) and Xavier (\ref{fig:gpu_untuned_xavier_mobilenetv1}-\ref{fig:gpu_untuned_xavier_vgg}) devices, with overall trends shown in Table~\ref{tab:cifar10_gpu_results}.
We note that the HiKey's inference time on the GPU is much higher than on the CPU, e.g., the dense \emph{direct} MobileNetV1 is almost 7$\times$ slower on the GPU.
For the pruned experiments, we see speedups most consistently using \emph{spatial pack}, which is different from the CPU where we almost always saw a slowdown.
In terms of expected speedup, for the HiKey and Xavier respectively, for weight pruning we achieve 7.4\% and 2.2\%, and for channel pruning we achieve 41.1\% and 26.0\%.
For \texttt{float16}, unlike the CPU, we observe speedups in several cases, however this behavior is not consistent across models, algorithms, or devices.
For example, on the HiKey using \emph{GEMM} with MobileNetV1 (Figure~\ref{fig:gpu_untuned_hikey_mobilenetv1}), \texttt{float16} provides a slowdown, whereas for other models on the HiKey we observe a speedup, as well as a speedup for the same model on the Xavier.
On the Xavier, \texttt{float16} provides a speedup in all cases except for \emph{direct} convolution where we observe small slowdowns.
On average \texttt{float16} achieves 51.9\% and 49.4\% of its potential speedup on the HiKey and Xavier respectively.

\begin{table}[]
\footnotesize
\caption{Analysis of CIFAR-10 GPU results for varying combinations of parameters, summarizing Figure~\ref{fig:inference_cifar10_gpu}.
The best inference times are shown in \textbf{bold}.
Shortened names are used for brevity: WP (weight pruning), CP (channel pruning), i8 (\texttt{int8}), f16 (\texttt{float16}).
\label{tab:cifar10_gpu_results}}
\begin{tabular}{cl|ccccc|ccc|c}
\toprule
\multirow{2}{*}{\textbf{Platform}} & \multirow{2}{*}{\textbf{Model}} & \multicolumn{5}{|c|}{\textbf{Fastest algorithm}}             & \multicolumn{3}{|c|}{\textbf{Fastest compression technique}} & \multirow{2}{*}{\shortstack{\textbf{Overall}\\\textbf{fastest}}} \\
                                   &                                 & Dense & WP            & CP   & i8              & f16    & GEMM              & Direct            & Spatial (Pack)       &                                   \\
                                   \hline
\multirow{4}{*}{\shortstack{HiKey\\(untuned)}} & MobileNetV2                     & GEMM     & Spatial & Spatial          & GEMM            & \textbf{Direct} & f16           & \textbf{f16}         & WP                  & f16+Direct                        \\
                                   & ResNet18                        & Spatial  & Spatial & Spatial          & \textbf{Direct} & GEMM            & f16           & \textbf{i8}          & CP                  & i8+Direct                         \\
                                   & VGG-16                          & Spatial  & Spatial & \textbf{Spatial} & GEMM            & GEMM            & i8            & i8                   & \textbf{CP}         & CP+Spatial                        \\
                                   & MobileNetV1                     & GEMM     & Spatial & Spatial          & \textbf{Direct} & Direct          & Dense         & i8                   & WP                  & i8+Direct                          \\
                                   \hline
\multirow{4}{*}{\shortstack{Xavier\\(untuned)}}                         & MobileNetV2                     & Direct   & Direct  & Direct           & \textbf{Direct} & Direct          & i8            & \textbf{i8}          & i8                  & i8+Direct                         \\
                                   & ResNet18                        & Direct   & Spatial & Spatial          & \textbf{Direct} & Direct          & i8            & \textbf{i8}          & i8                  & i8+Direct                         \\
                                   & VGG-16                          & Direct   & Spatial & Direct           & \textbf{Direct} & Direct          & i8            & \textbf{i8}          & i8                  & i8+Direct                         \\
                                   & MobileNetV1                     & Direct   & Direct  & Direct           & \textbf{Direct} & Direct          & i8            & \textbf{i8}          & i8                  & i8+Direct
  \\
  \hline
\multirow{4}{*}{\shortstack{HiKey\\(tuned)}}             & MobileNetV2                     & Direct   & -  & -  & Direct           & \textbf{GEMM}   & \textbf{f16}       & Dense             & f16               & f16+GEMM                          \\
                                   & ResNet18                        & Direct   & -  & -  & Spatial          & \textbf{Direct} & i8                 & \textbf{f16}      & f16               & f16+Direct                        \\
                                   & VGG-16                          & GEMM     & -  & -  & \textbf{Direct}  & Spatial         & Dense              & \textbf{i8}       & f16               & i8+Direct                         \\
                                   & MobileNetV1                     & Direct   & -  & -  & \textbf{Spatial} & Direct          & i8                 & i8                & \textbf{i8}       & i8+Spatial
                                   \\ \hline
\multirow{4}{*}{\shortstack{Xavier\\(tuned)}}            & MobileNetV2                     & Direct   & -  & -  & \textbf{Direct}  & Direct          & i8                 & \textbf{i8}       & i8                & i8+Direct                         \\
                                   & ResNet18                        & Direct   & -  & -  & \textbf{Direct}  & Direct          & i8                 & \textbf{i8}       & i8                & i8+Direct                         \\
                                   & VGG-16                          & Direct   & -  & -  & \textbf{Direct}  & Direct          & i8                 & \textbf{i8}       & i8                & i8+Direct                         \\
                                   & MobileNetV1                     & Direct   & -  & -  & \textbf{Direct}  & Direct          & i8                 & \textbf{i8}       & i8                & i8+Direct         \\
                                   \bottomrule
\end{tabular}
\end{table}

\subsubsection{Inference -- GPU (tuned)}

The last two rows of Figure~\ref{fig:inference_cifar10_gpu} show the tuned performance of the CIFAR-10 models on GPUs, with overall trends shown in Table~\ref{tab:cifar10_gpu_results}.
As noted in Section~\ref{subsec:setup:sys_software}, we cannot provide tuned results for sparse models on the GPU.
The HiKey GPU is still slower than the HiKey CPU (tuned), with the best dense result being 3.1$\times$ slower on average.
The Xavier does not get any improvement when tuning, we discuss this issue in Section~\ref{subsec:discuss:sys_software}.
For quantization, on the HiKey, taking the best result for each model, we achieve 58.9\% and 44.4\% of the expected speedup on average for \texttt{float16} and \texttt{int8} respectively; the Xavier achieves 49.0\% and 28.4\% of its expected speedups.

\subsection{ImageNet}
\label{subsec:eval:imagenet}

\subsubsection{Accuracy}

For the four models, the baseline (dense) top-1 accuracy on ImageNet is shown in Table~\ref{tab:acc_imagenet}.
EfficientNetB0 has the highest accuracy, which may be surprising given it has fewer parameters.
However, EfficientNetB0 is more recent and thus exploits a number of newer machine learning techniques to improve its parameter and training efficiency.

The accuracy on ImageNet with varying levels of compression can be seen in the second row of Figure~\ref{fig:accuracies}.
We observe a similar trend to the CIFAR-10 models, namely the smaller models (EfficientNetB0 and MobileNetV2) lose their accuracy more quickly than the larger ones (DenseNet161 and ResNet50).
We also observe that \emph{all} models lose more accuracy earlier when compared to CIFAR-10 pruning.
This suggests that the CIFAR-10 models are more overparameterized.

For data-type quantization, we observe a similar trend as CIFAR-10, namely a negligible difference in accuracy for \texttt{float16}.
For ResNet50, we see a large drop in accuracy for uncalibrated \texttt{int8} quantization, recovering to around a 3.0\% accuracy reduction.
For MobileNetV2, we observe a huge drop in accuracy for the uncalibrated model, down to around 0.09\%, recovering to around a 6.6\% accuracy reduction.
This drop was much higher than we expected, so we also tried importing the Keras~\cite{cholletKeras2015} definition of MobileNetV2, and observed the same behavior.

For EfficientNetB0, we also observe a huge drop in accuracy to 0.08\%, however the recovery is much smaller than MobileNetV2's, reaching only 0.43\% accuracy.
This is due to architecture features of the model which make it less suitable for quantization, which we discuss in Section~\ref{subsub:discuss:models}.
For DenseNet161, we cannot run the \texttt{int8} model in TVM due to an unsupported quantized operation.
This excludes it from collection of uncalibrated accuracy and inference time results.
However, when calibrated in ONNXRuntime we reduce accuracy by 1.9\%.

\begin{table}[t]
\centering
\footnotesize
\caption{
  ImageNet models, including baseline accuracy (Top1), and our chosen compression ratios and corresponding accuracies.
  \label{tab:acc_imagenet}
}
\begin{tabular}{lccccccc}
\toprule
\multirow{2}{*}{Model} & \multirow{2}{*}{Params} & \multirow{2}{*}{MACs} & \multirow{2}{*}{Top1} & \multicolumn{4}{c}{Model Optimization Accuracy (\& Compression Ratio)}      \\ \cline{5-8}
                       &                         &                       &                       & Weight Pruning  & Channel Pruning  & \texttt{float16} & \texttt{int8} \\ \midrule
MobileNetV2            & 3.5M                    & 327M                  & 71.9\%                & 58.4\% (80\%)   & 49.9\% (50\%) & 71.9\% (50\%) & 65.3\% (75\%)     \\
ResNet50               & 25.6M                   & 4.1G                  & 76.1\%                & 67.3\% (95\%)   & 46.6 (80\%)   & 76.1\% (50\%) & 73.1\% (75\%)   \\
DenseNet161            & 27.7M                   & 7.8G                  & 77.1\%                & 74.3\% (95\%)   & 62.7\ (80\%)  & 77.1\% (50\%) & 75.2\% (75\%) \\
EfficientNetB0          & 5.3M                    & 415M                  & 77.7\%                & 65.3\% (80\%)   & 54.9 (50\%)  & 77.6\% (50\%) & 0.4\% (75\%)   \\
\bottomrule
\end{tabular}
\end{table}

\subsubsection{Inference -- CPU (untuned)}

\begin{figure*}[t]
  \centering
  \begin{subfigure}{0.95\textwidth}
    \centering     \hspace*{0.2cm}    \includegraphics[width=\textwidth]{fig_neo/inference/legend.png}
  \end{subfigure}%
  \\
\centering
    \begin{subfigure}{0.24\textwidth}
    \centering     \includegraphics[width=\textwidth]{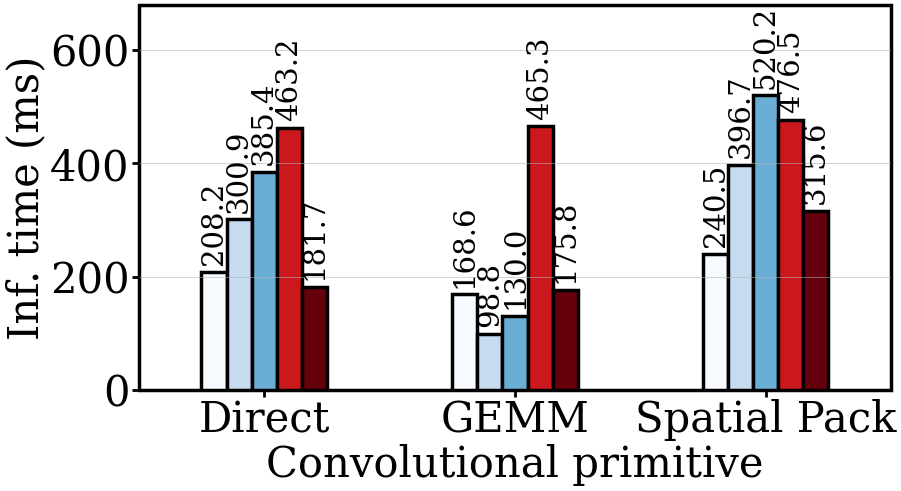}
    \caption{\footnotesize EfficientNetB0 on HiKey} \label{fig:imagenet_hikey_EfficentNetB0}
  \end{subfigure}%
  \hspace*{\fill} 
  \begin{subfigure}{0.24\textwidth}
    \centering     \includegraphics[width=\textwidth]{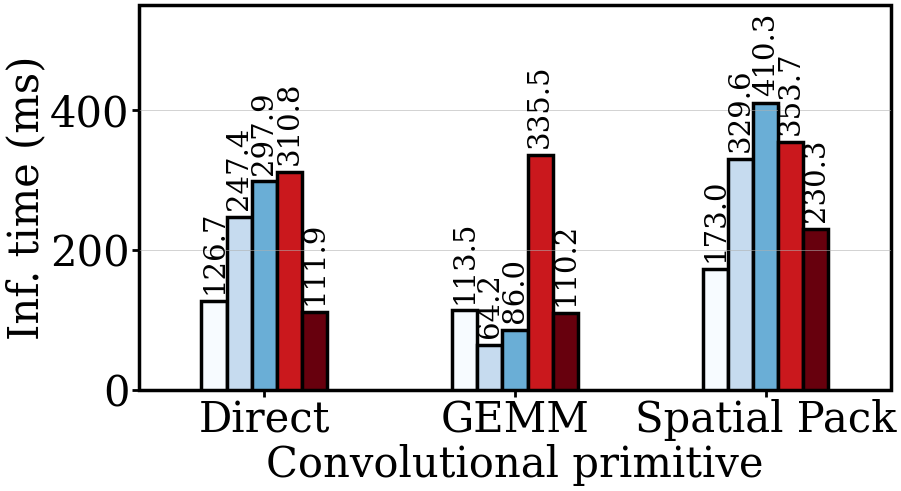}
    \caption{\footnotesize MobileNetV2 on HiKey} \label{fig:imagenet_hikey_mobilenetv2}
  \end{subfigure}%
  \hspace*{\fill}
  \begin{subfigure}{0.24\textwidth}
    \centering     \includegraphics[width=\textwidth]{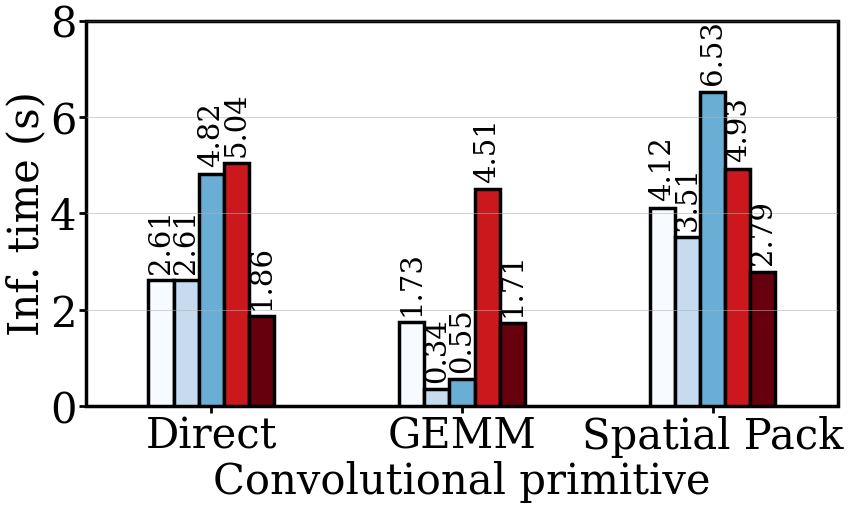}
    \caption{\footnotesize ResNet50 on HiKey} \label{fig:imagenet_hikey_resnet50}
  \end{subfigure}%
  \hspace*{\fill} 
  \begin{subfigure}{0.24\textwidth}
    \centering     \includegraphics[width=\textwidth]{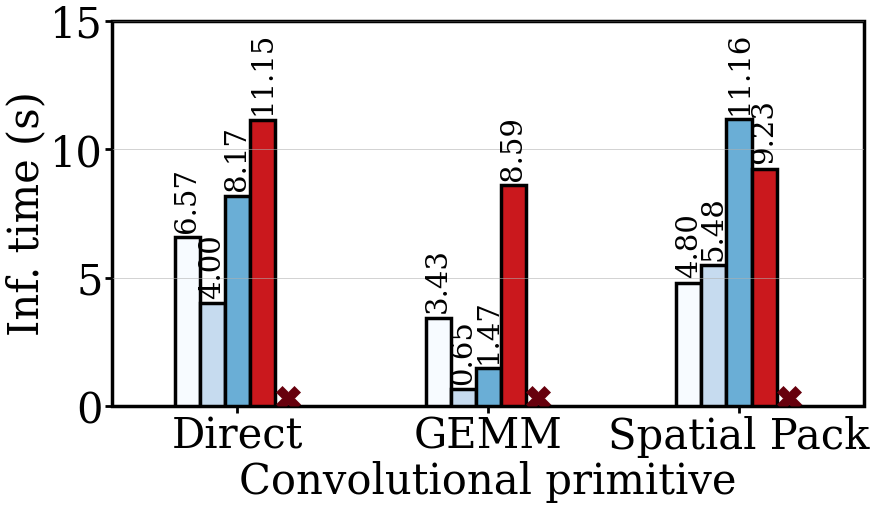}
    \caption{\footnotesize DenseNet161 on HiKey} \label{fig:imagenet_hikey_densenet}
  \end{subfigure}
  \\
    \begin{subfigure}{0.24\textwidth}
    \centering     \includegraphics[width=\textwidth]{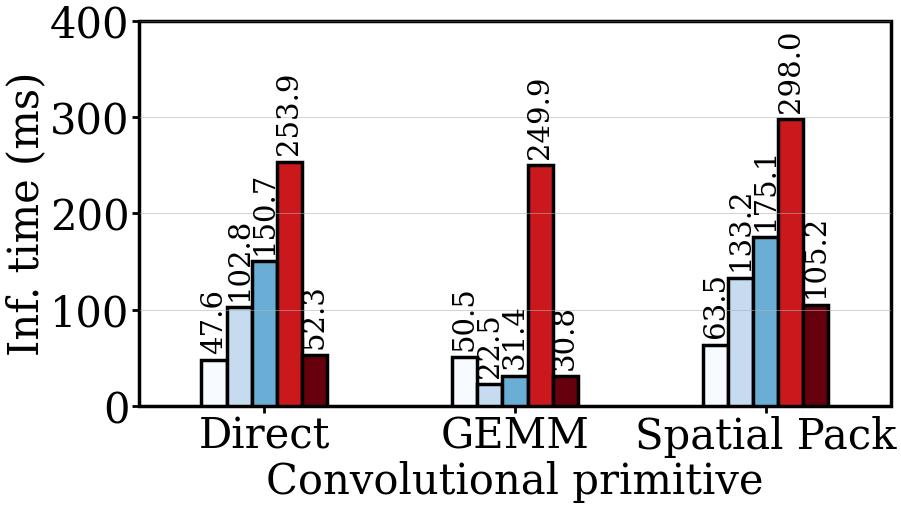}
    \caption{\footnotesize EfficientNetB0 on Intel i7} \label{fig:imagenet_i7_EfficentNetB0}
  \end{subfigure}
    \hspace*{\fill} 
 \begin{subfigure}{0.24\textwidth}
    \centering     \includegraphics[width=\textwidth]{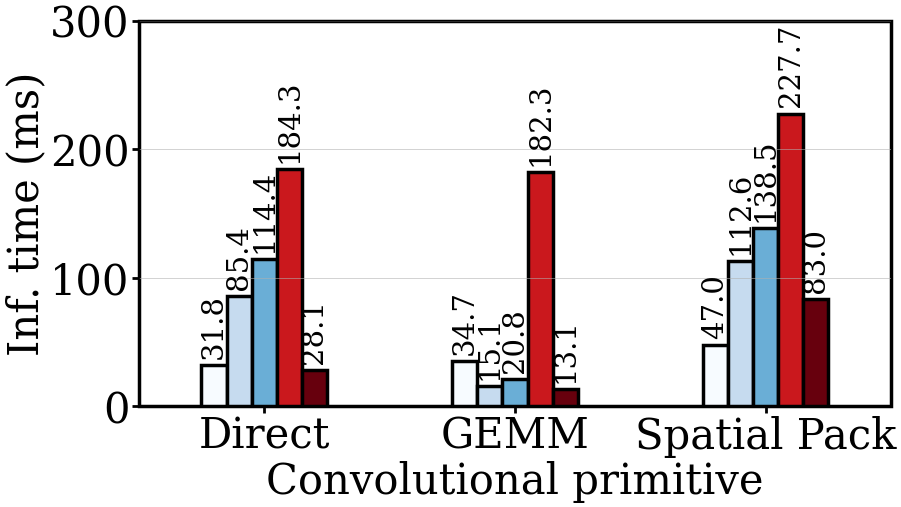}
    \caption{\footnotesize MobileNetV2 on Intel i7} \label{fig:imagenet_i7_mobilenetv2}
  \end{subfigure}
  \hspace*{\fill}
  \begin{subfigure}{0.24\textwidth}
    \centering     \includegraphics[width=\textwidth]{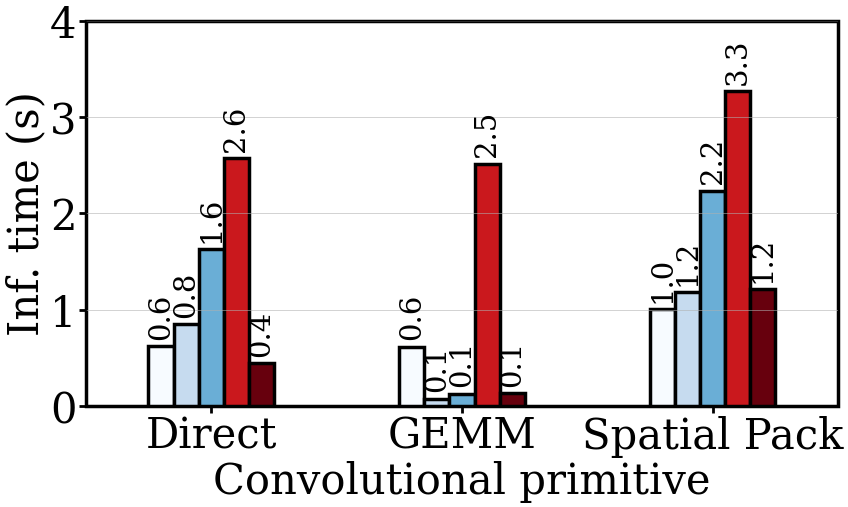}
    \caption{\footnotesize ResNet50 on Intel i7} \label{fig:imagenet_i7_resnet50}
  \end{subfigure}%
  \hspace*{\fill} 
  \begin{subfigure}{0.24\textwidth}
    \centering     \includegraphics[width=\textwidth]{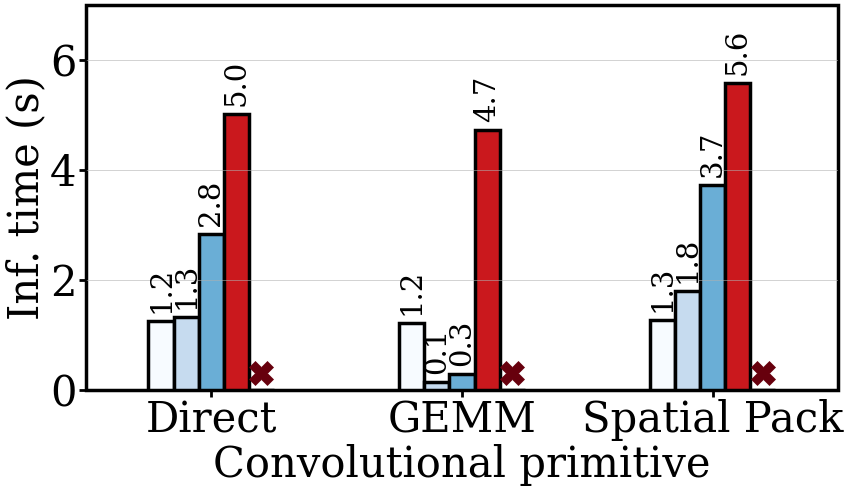}
    \caption{\footnotesize DenseNet161 on Intel i7} \label{fig:imagenet_i7_densenet}
  \end{subfigure}
  \\
    \begin{subfigure}{0.24\textwidth}
    \centering     \includegraphics[width=\textwidth]{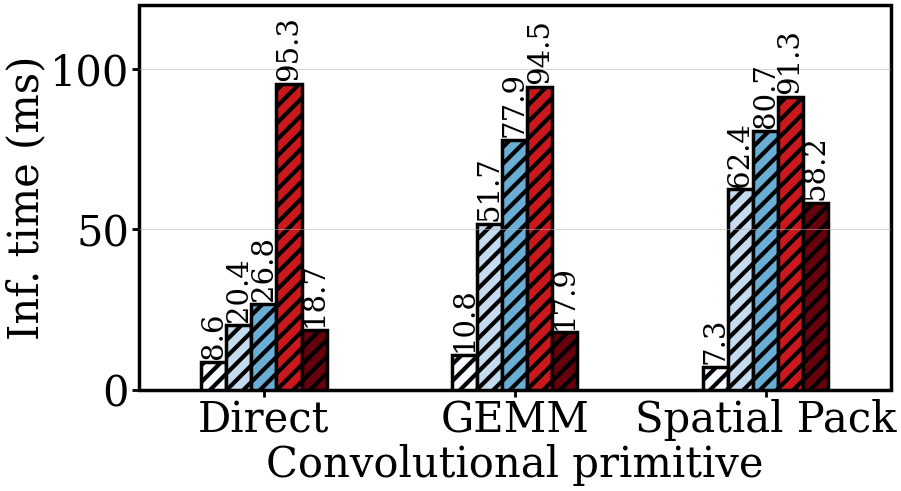}
    \caption{\footnotesize EfficientNetB0 on Intel i7} \label{fig:cpu_tuned_imagenet_i7_EfficentNetB0}
  \end{subfigure}
  \hspace*{\fill} 
    \begin{subfigure}{0.24\textwidth}
    \centering     \includegraphics[width=\textwidth]{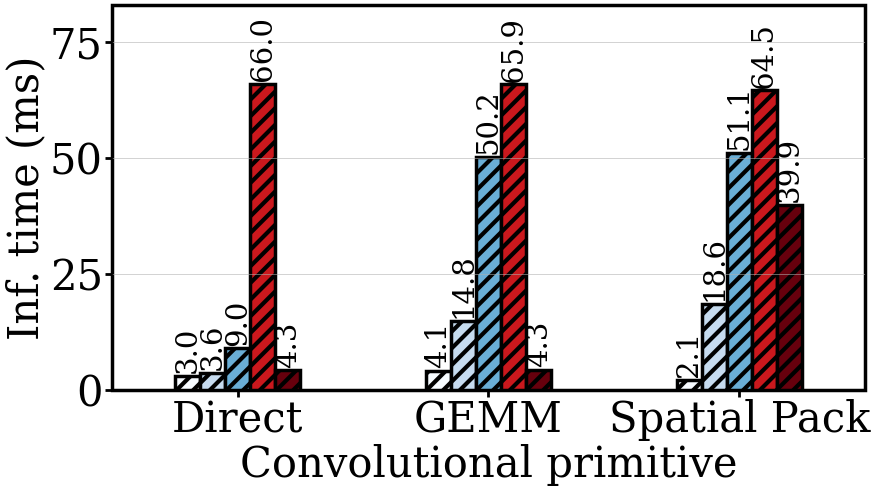}
    \caption{\footnotesize MobileNetV2 on Intel i7} \label{fig:cpu_tuned_imagenet_i7_mobilenetv2}
  \end{subfigure}
  \hspace*{\fill} 
    \begin{subfigure}{0.24\textwidth}
    \centering     \includegraphics[width=\textwidth]{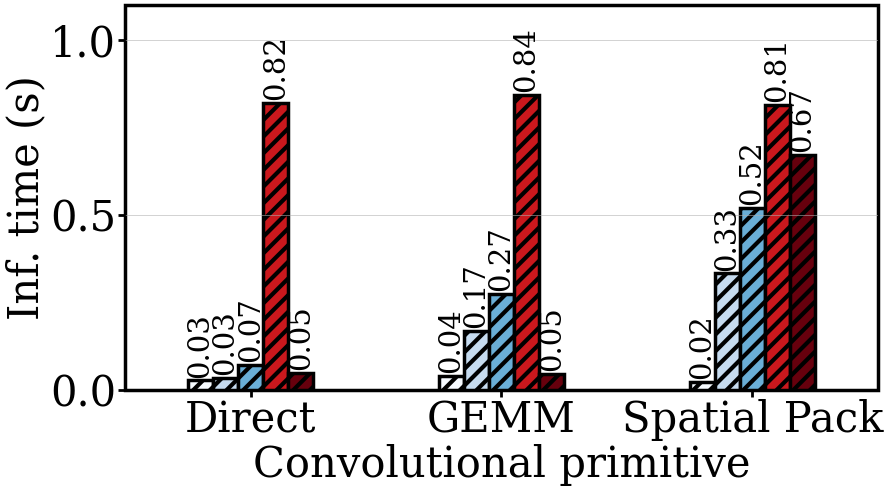}
    \caption{\footnotesize ResNet50 on Intel i7} \label{fig:cpu_tuned_imagenet_i7_resnet50}
  \end{subfigure}%
    \hspace*{\fill} 
    \begin{subfigure}{0.24\textwidth}
    \centering     \includegraphics[width=\textwidth]{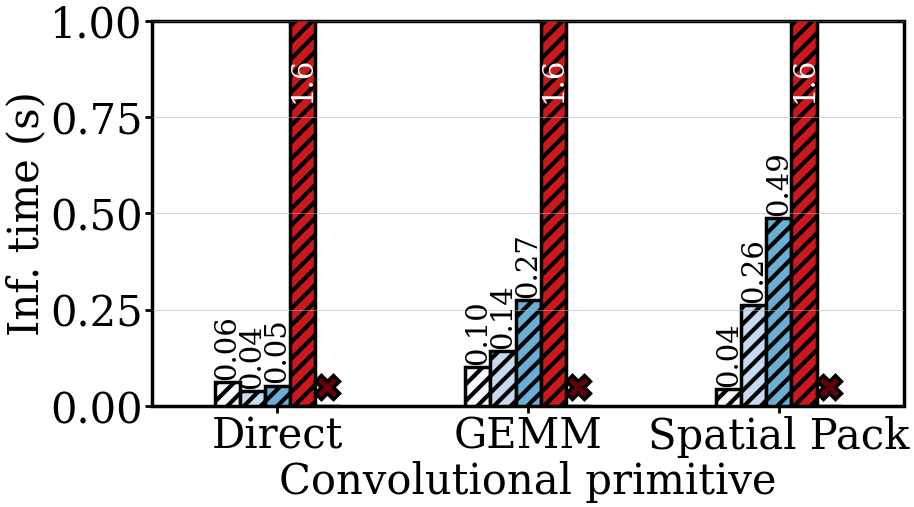}
    \caption{\footnotesize DenseNet161 on Intel i7} \label{fig:cpu_tuned_imagenet_i7_densenet}
  \end{subfigure}%
  \\
    \caption{ Experiments comparing the compressed ImageNet models chosen from obvious elbows of accuracy, with varying algorithmic primitives, benchmarked on the i7 and HiKey CPU platforms, with and without auto-scheduling.}
  	\label{fig:inference_imagenet_cpu}
\end{figure*}

The first two rows of Figure~\ref{fig:inference_imagenet_cpu} show the untuned performance of the ImageNet models when running on the i7 and HiKey CPUs, with overall trends shown in Table~\ref{tab:imagenet_results}.
For dense models, we observe that in all cases on the HiKey \emph{GEMM} gives the best performance, which matches its behavior as seen for CIFAR-10.
For the i7, \emph{GEMM} is fastest for the large models (ResNet50 and DenseNet161), however \emph{direct} is fastest for the small models (MobileNetV2 and EfficientNetB0); on CIFAR-10 \emph{direct} was consistently the fastest on this CPU.

For weight pruning, we find that by taking the best performing variants as before, we achieve 30.3\% and 41.8\% of the potential speedup for the HiKey and i7 respectively; significantly higher than CIFAR-10.
For channel pruning, this is 60.2\% and 84.2\% respectively, which is 16.7\% less than CIFAR-10 for the HiKey and 0.3\% more for the i7.
For quantization, we see similar trends to CIFAR-10, namely a slowdown using \texttt{float16} and a speedup using \texttt{int8}.
For \texttt{int8}, we achieve 25.0\% and 73.0\% of the expected speedup on the i7 and HiKey respectively, lower than CIFAR-10.

\begin{table}[]
\footnotesize
\caption{Analysis of ImageNet results for varying combinations of parameters, summarizing Figures~\ref{fig:inference_imagenet_cpu} and~\ref{fig:inference_imagenet_gpu}.
The best inference times are shown in \textbf{bold}.
Shortened names are used for brevity: WP (weight pruning), CP (channel pruning), i8 (\texttt{int8}), f16 (\texttt{float16}).
\label{tab:imagenet_results}}
\begin{tabular}{cl|ccccc|ccc|c}
\toprule
\multirow{2}{*}{\textbf{Platform}} & \multirow{2}{*}{\textbf{Model}} & \multicolumn{5}{|c|}{\textbf{Fastest algorithm}}             & \multicolumn{3}{|c|}{\textbf{Fastest compression technique}} & \multirow{2}{*}{\shortstack{\textbf{Overall}\\\textbf{fastest}}} \\
                                   &                                 & Dense & WP            & CP   & i8              & f16    & GEMM              & Direct            & Spatial (Pack)       &                                   \\
                                   \hline
\multirow{4}{*}{\shortstack{HiKey CPU \\ (untuned)}}             & MobileNetV2                     & GEMM     & \textbf{GEMM} & GEMM & GEMM          & Direct & \textbf{WP}         & i8             & Dense               & WP+GEMM                           \\
                                   & ResNet50                        & GEMM     & \textbf{GEMM} & GEMM & GEMM          & GEMM   & \textbf{WP}         & i8             & i8                  & WP+GEMM                           \\
                                   & DenseNet161                     & GEMM     & \textbf{GEMM} & GEMM & -          & GEMM   & \textbf{WP}         & WP             & Dense               & WP+GEMM                           \\
                                   & EfficientNetB0                  & GEMM     & \textbf{GEMM} & GEMM & GEMM          & Direct & \textbf{WP}         & i8             & Dense               & WP+GEMM
                                   \\\hline
\multirow{4}{*}{\shortstack{i7\\(untuned)}}                & MobileNetV2                     & Direct   & GEMM          & GEMM & \textbf{GEMM} & GEMM   & \textbf{i8}         & i8             & Dense               & i8+GEMM                           \\
                                   & ResNet50                        & GEMM     & \textbf{GEMM} & GEMM & GEMM          & GEMM   & \textbf{WP}         & i8             & Dense               & WP+GEMM                           \\
                                   & DenseNet161                     & GEMM     & \textbf{GEMM} & GEMM & -             & GEMM   & \textbf{WP}         & Dense          & Dense               & WP+GEMM                           \\
                                   & EfficientNetB0                  & Direct   & \textbf{GEMM} & GEMM & GEMM          & GEMM   & \textbf{WP}         & Dense           & Dense               & WP+GEMM
                                   \\ \hline
\multirow{4}{*}{\shortstack{i7\\(tuned)}}             & MobileNetV2                     & \textbf{Spatial} & Direct          & Direct & Direct & Spatial & Dense         & Dense              & \textbf{Dense}        & Dense+Spatial                     \\
                                   & ResNet50                        & \textbf{Spatial} & Direct          & Direct & GEMM   & Spatial & Dense         & Dense              & \textbf{Dense}        & Dense+Spatial                     \\
                                   & DenseNet161                     & Spatial          & \textbf{Direct} & Direct & -      & Spatial & Dense         & \textbf{WP}        & Dense                 & WP+Direct                         \\
                                   & EfficientNetB0                  & \textbf{Spatial} & Direct          & Direct & GEMM   & Spatial & Dense         & Dense              & \textbf{Dense}        & Dense+Spatial                     \\
                                   \hline\hline
\multirow{4}{*}{\shortstack{HiKey GPU\\(untuned)}}             & MobileNetV2                     & GEMM     & Spatial          & Direct  & Direct          & \textbf{GEMM}   & \textbf{f16}       & i8                & i8                & f16+GEMM                          \\
                                   & ResNet50                        & Direct   & Spatial          & Spatial & Direct          & Direct          & WP                 & i8                & \textbf{WP}       & WP+Spatial                        \\
                                   & DenseNet161                     & Direct   & \textbf{Spatial} & Spatial & -               & GEMM            & WP                 & WP                & \textbf{WP}       & WP+Spatial                        \\
                                   & EfficientNetB0                  & GEMM     & Spatial          & Spatial & \textbf{Direct} & GEMM            & f16                & \textbf{i8}       & i8                & i8+Direct                         \\
  \hline
\multirow{4}{*}{\shortstack{Xavier\\(untuned)}}            & MobileNetV2                     & Direct   & Spatial          & Spatial & Direct          & \textbf{Direct} & f16                & \textbf{f16}      & f16               & f16+Direct                        \\
                                   & ResNet50                        & Spatial  & Spatial          & Spatial & \textbf{Direct} & Direct          & i8                 & \textbf{i8}       & i8                & i8+Direct                         \\
                                   & DenseNet161                     & Spatial  & \textbf{Spatial} & Spatial & -               & Direct          & WP                 & f16               & \textbf{WP}       & WP+Spatial                        \\
                                   & EfficientNetB0                  & Direct   & Spatial          & Spatial & Direct          & \textbf{Direct} & f16                & \textbf{f16}      & f16               & f16+Direct                          \\
                                   \bottomrule
\end{tabular}
\end{table}

\subsubsection{Inference -- CPU (tuned)}
\label{subsubsec:eval_cpu_inf_imagenet_tuned}

The last row of Figure~\ref{fig:inference_imagenet_cpu} shows the tuned performance of the ImageNet models when running on the i7 CPU.
We note that tuning on the HiKey CPU (and GPU) was not practical, since the two variants we attempted took over 140 hours each, so we do not include any of the 57 variants required for each device.
For the dense case on the i7, we see that \emph{spatial pack} is consistently the best, matching the observed trends on tuned CIFAR-10.
For the pruned models, we do not see any cases where pruned models are faster than a dense \texttt{float32} implementation.
This is contrasted with CIFAR-10, where we observe this in every case.

\subsubsection{Inference -- GPU (untuned)}
\begin{figure*}[t]
  \centering
  \begin{subfigure}{0.95\textwidth}
    \centering      \hspace*{0.2cm}   \includegraphics[width=\textwidth]{fig_neo/inference/legend.png}
  \end{subfigure}%
  \\
\centering
    \begin{subfigure}{0.24\textwidth}
    \centering     \includegraphics[width=\textwidth]{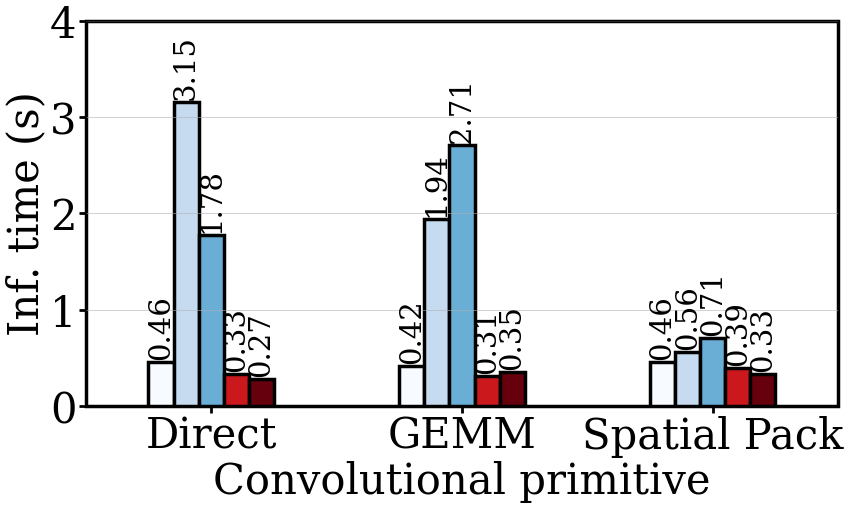}
    \caption{\footnotesize EfficientNetB0 on HiKey} \label{fig:gpu_imagenet_hikey_EfficentNetB0}
  \end{subfigure}%
  \hspace*{\fill}
  \begin{subfigure}{0.24\textwidth}
    \centering     \includegraphics[width=\textwidth]{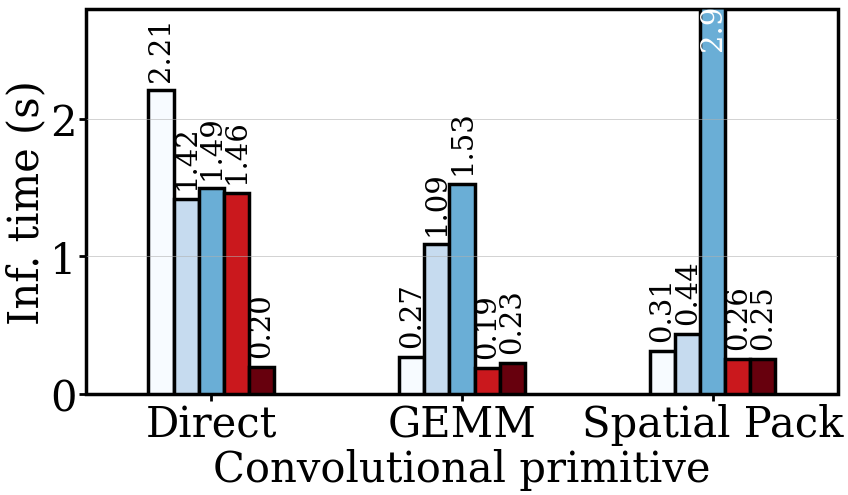}
    \caption{\footnotesize MobileNetV2 on HiKey} \label{fig:gpu_imagenet_hikey_mobilenetv2}
  \end{subfigure}%
  \hspace*{\fill}
  \begin{subfigure}{0.24\textwidth}
    \centering     \includegraphics[width=\textwidth]{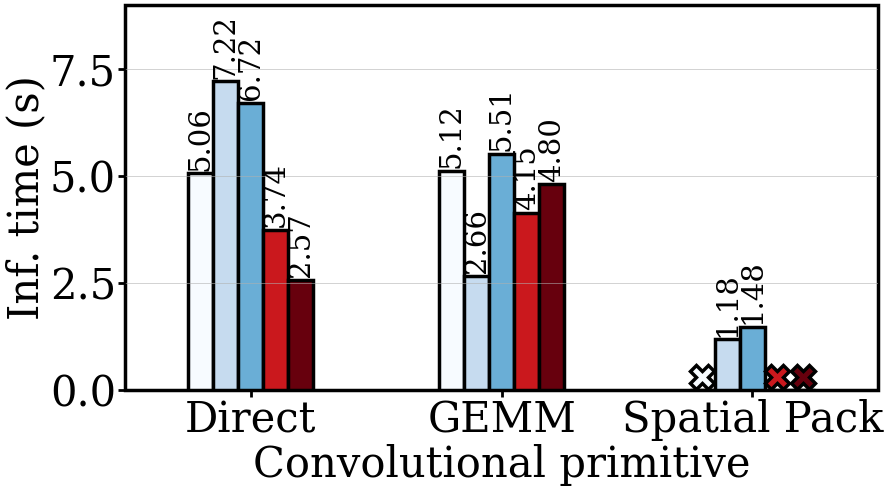}
    \caption{\footnotesize ResNet50 on HiKey} \label{fig:gpu_imagenet_hikey_resnet50}
  \end{subfigure}%
  \hspace*{\fill} 
  \begin{subfigure}{0.24\textwidth}
    \centering     \includegraphics[width=\textwidth]{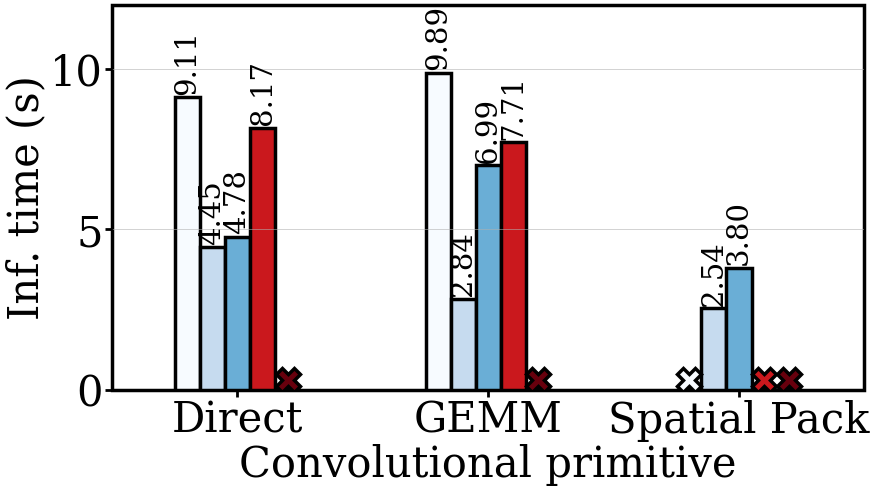}
    \caption{\footnotesize DenseNet161 on HiKey} \label{fig:gpu_imagenet_hikey_densenet}
  \end{subfigure}
  \\

    \begin{subfigure}{0.24\textwidth}
    \centering     \includegraphics[width=\textwidth]{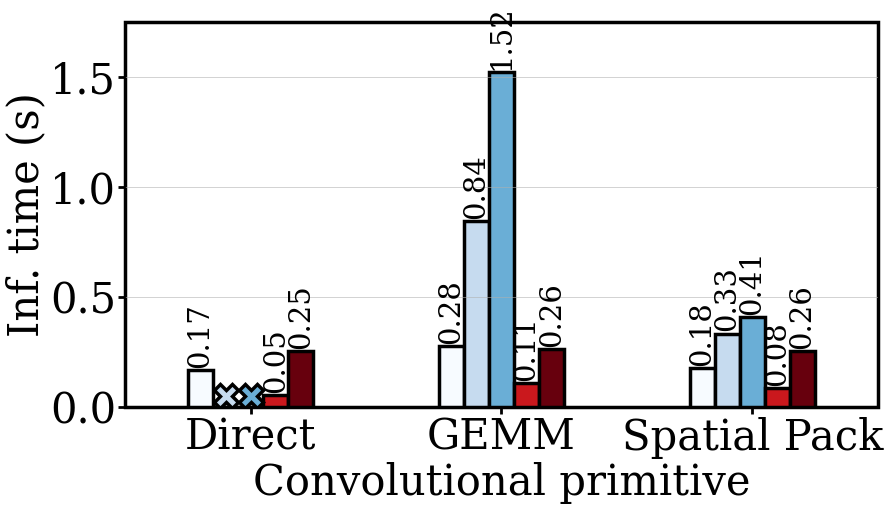}
    \caption{\footnotesize EfficientNetB0 on Xavier} \label{fig:gpu_imagenet_xavier_EfficentNetB0}
  \end{subfigure}
    \hspace*{\fill}
    \begin{subfigure}{0.24\textwidth}
    \centering     \includegraphics[width=\textwidth]{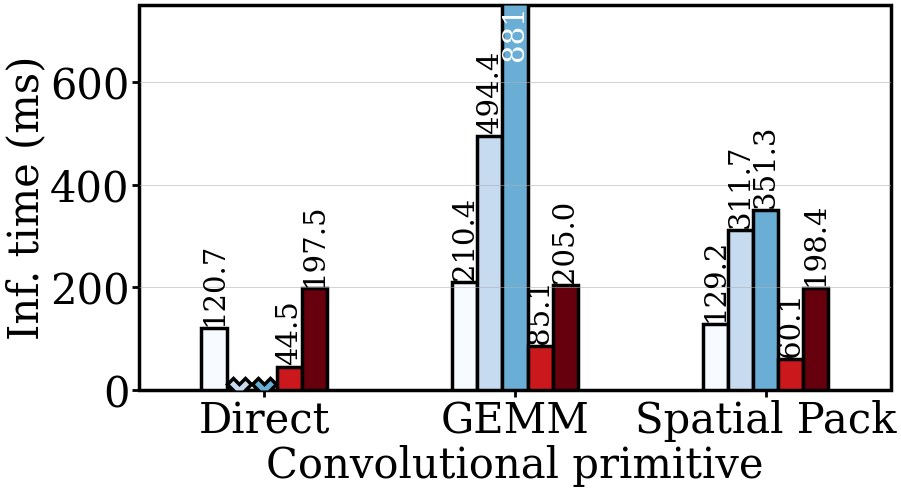}
    \caption{\footnotesize MobileNetV2 on Xavier} \label{fig:gpu_imagenet_xavier_mobilenetv2}
  \end{subfigure}
  \hspace*{\fill}
  \begin{subfigure}{0.24\textwidth}
    \centering     \includegraphics[width=\textwidth]{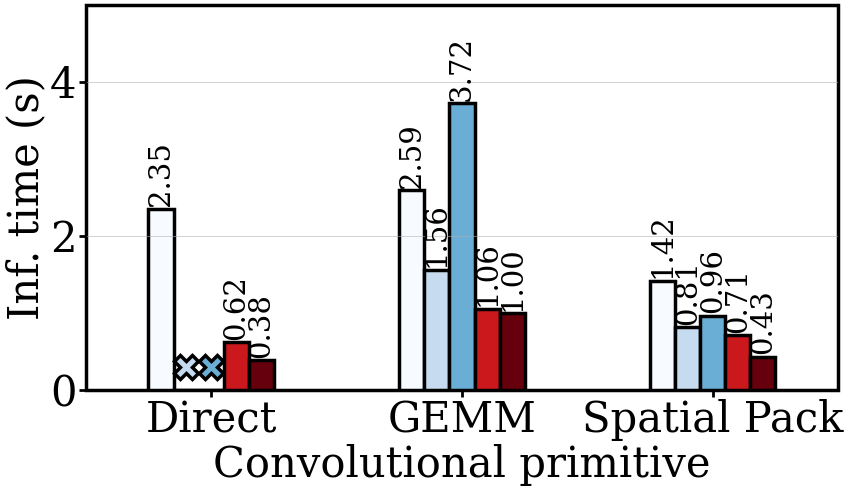}
    \caption{\footnotesize ResNet50 on Xavier} \label{fig:gpu_imagenet_xavier_resnet50}
  \end{subfigure}%
  \hspace*{\fill}
  \begin{subfigure}{0.24\textwidth}
    \centering     \includegraphics[width=\textwidth]{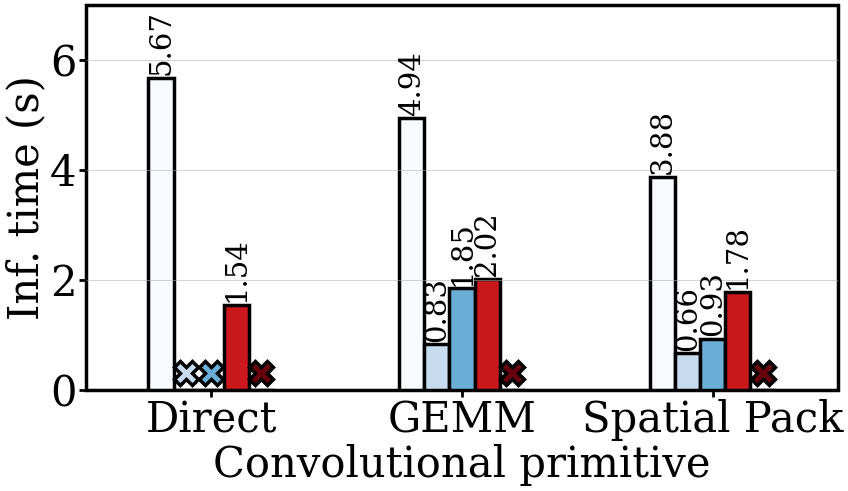}
    \caption{\footnotesize DenseNet161 on Xavier} \label{fig:gpu_imagenet_xavier_densenet}
  \end{subfigure}%
    \caption{ Experiments comparing the compressed ImageNet models chosen from obvious elbows of accuracy, with varying algorithmic primitives, benchmarked on the HiKey and Xavier GPU platforms, without auto-scheduling.}
  	\label{fig:inference_imagenet_gpu}
\end{figure*}

On the Xavier in the dense case, \emph{spatial pack} is the best algorithm for the larger models (ResNet50 and DenseNet161), and \emph{direct} is the best for the smaller models (MobileNetV2 and EfficientNetB0).
On the Hikey, for the smaller models \emph{GEMM} was the best, and for the larger models \emph{direct} was the best.
However, on the HiKey, dense ResNet50 and DenseNet161 experiments using \emph{spatial pack} crashed with the error \texttt{CL\_INVALID\_WORK\_GROUP\_SIZE}.
This means that TVM is exceeding the number of supported work items (see the OpenCL specification for more details~\cite{stoneOpenCLParallelProgramming2010}).
If we run auto-tuning, TVM can configure the work group size, which could avoid this issue.

For the sparse experiments, \emph{spatial pack} using weight pruning was consistently the best across both GPUs, however only outperformed the baseline in one case, ResNet50 on the Xavier.
On the Xavier, we found that sparse \emph{direct} experiments did not halt, even allowing hours for a single run.
GPU memory utilization was at its maximum, suggesting that some inefficiency in this algorithm/hardware combination.
Again, auto-tuning may make this variant viable, however as highlighted in Section~\ref{subsec:setup:sys_software}, we cannot tune sparse models on GPUs.

For ImageNet, quantization was not consistently the best compression technique for the GPUs, unlike CIFAR-10.
For several cases, weight pruning performed best.
On the HiKey, using \texttt{int8} for EfficientNet was the fastest approach, however it should be noted that in this case the accuracy drop was prohibitively large, as discussed in Section~\ref{subsub:discuss:models}.

\subsubsection{Inference -- GPU (tuned)}

As discussed in Section~\ref{subsubsec:eval_cpu_inf_imagenet_tuned}, collecting tuned results for the HiKey GPU was not practical.
In addition, we again observed no speedup on the Xavier when tuning, hence we do not include the graphs.

\section{Discussion and Related Work}
\label{sec:discussion}

In Section~\ref{sec:evaluation} we observed many variations across our experiments with a number of non-trivial dynamics emerging.
Our main observation is \textit{I) MACs, accuracy, and inference time are not strongly correlated}, along with 12 additional observations, \textit{II)-XIII)}.
These observations are not intended to be definitive, and large changes in the results could be expected by bringing in new techniques from DLAS.
However, the key point of this work is that across-stack interactions of machine learning and systems optimizations can be non-trivial, and bringing in additional features may significantly accelerate or impede a given technique.
We refer the reader to other characterization works and surveys which highlight cutting edge techniques from across DLAS~\cite{szeEfficientProcessingDeep2020,blalockWhatStateNeural2020,hadidiCharacterizingDeploymentDeep2019,hookerHardwareLottery2021,liDeepLearningCompiler2021,wangConvergenceEdgeComputing2020}.

\subsection{Datasets \& Problem Spaces}

Between datasets, models showed similar trends in terms of accuracy, with the accuracy losses due to compression being higher for ImageNet models.
The computational requirements of models for each dataset varied, i.e., CIFAR-10 models have fewer MACs and parameters than ImageNet models (Tables~\ref{tab:acc_cifar10} and~\ref{tab:acc_imagenet}), which we would expect to have effects on the inference behavior.
For instance, \textit{II) tuned sparse CIFAR-10 models were faster than the dense baseline, but this was not the case for ImageNet models.}
Possible explanations include overheads in the sparse algorithm and data format choices, which could be exacerbated by the larger ImageNet models; and for tuning, both sets of models were given the same number of trials despite ImageNet models being larger.

\subsection{Models \& Neural Architectures}
\label{subsub:discuss:models}

As a related observation to \textit{I)}, we note that
\textit{III) model size is not strongly correlated with accuracy, however smaller models are more vulnerable to compression.}
For example, EfficientNetB0 has a higher baseline accuracy than both DenseNet161 and ResNet50, despite having at most 21\% of the parameters.
This can be understood as EfficientNetB0 being a more recent model which exploits novel architectural and training techniques to achieve better parameter efficiency.
However, other works have shown that if we keep the model architecture the same, more parameters generally means higher accuracy~\cite{crowleyMoonshineDistillingCheap2018,gibsonOptimizingGroupedConvolutions2020,tanEfficientNetRethinkingModel2019}.
For CIFAR-10, we observed that ResNet18 and MobileNetV2 had higher baseline accuracies than VGG-16, despite both having fewer parameters.
A similar explanation of using techniques such as residual blocks can explain this behavior.
However, as \textit{III)} notes, models with fewer parameters were more vulnerable to compression.

Another related observation is that EfficientNetB0 had the highest accuracy drops for \texttt{int8} quantization out of any model.
We understand this to be due to EfficientNetB0's architecture not being amenable to post-training quantization.
These issues were highlighted and corrected by EfficientNet-Lite~\cite{liuHigherAccuracyVision2020}, which removes the squeeze-and-excitation networks, and replaces the swish activation functions with ReLU6 activations~\cite{howardMobileNetsEfficientConvolutional2017}.

\subsection{Model Optimizations}

Overall, the best model optimization technique varied.
\textit{IV) Weight pruning tended to give better compression ratios and speedups than channel pruning, however achieved less of its expected speedups.}
We also observed that \textit{V) quantization's speedup varied by hardware platform and data-type.}

\subsubsection{Pruning}
\label{subsec:discuss:pruning}

As observation \textit{IV)} notes, weight pruning was generally faster than channel pruning, however the former achieved a lower proportion of its expected speedups.
There were several cases where a less compressed channel pruning model was faster than a more compressed weight pruning model.
However, this appears to come algorithmic interactions, since taking the best variant across algorithms it was rare for channel pruning to outperform weight pruning.
Overall, the relative performance of pruned models on the GPUs was worse than on the CPUs.
This is because sparse computations are irregular and thus cannot easily take full advantage of the greater number of cores available in a GPU that dense computations can.
However, some across-stack techniques can improve sparse performance on GPUs~\cite{guoAcceleratingSparseDNN2020,raduPerformanceAwareConvolutional2019}.
Since we could not tuned sparse computations on the GPUs, we cannot comment how well they would perform if the code further optimized.
Other pruning techniques that we did not explore include layer-wise pruning and gradient based methods, trade-offs of which are discussed in Blalock et al.~\cite{blalockWhatStateNeural2020}.
We only evaluated pruning techniques in terms of accuracy and compression, however as other work has noted, pruning also comes with non-negligible costs due to retraining, which may be a bottleneck to DSE~\cite{persandTaxonomySaliencyMetrics2021,huICEPickIterativeCostEfficient2023}.

\subsubsection{Data-type Quantization}
\label{subsubsec:discuss_quant}

\textit{VI) \texttt{float16} had a negligible impact on accuracy, when calibrated \texttt{int8} lost 1.8\% and 22.2\% for CIFAR-10 and ImageNet models on average respectively.}
Excluding EfficientNetB0, ImageNet models lost 3.8\% of accuracy on average.
With regards to inference time, as observation \textit{V)} notes for \texttt{float16} on the CPUs, we observed a consistent slowdown when compared to \texttt{float32}.
This is because the hardware runs \texttt{float16} computation using software emulation.
Although there are savings in memory footprint, the overhead involved in the emulation clearly outweighs these savings, with the differences exacerbated when tuning.
For \texttt{int8} on the CPUs, we generally observed a speedup, since the CPU ISAs can use SIMD instructions to compute \texttt{int8} computations relatively efficiently.

In the untuned \texttt{int8} case on the CPU, we observe some cases where we get \textit{higher} than expected speedups, for instance many of the \emph{GEMM} and \emph{direct} cases on the i7 in Figure~\ref{fig:inference_cifar10_cpu}.
The highest of this is for VGG-16 with CIFAR-10 using \emph{GEMM}, where we observe a speedup of over 9.2$\times$.
We hypothesize that using smaller data sizes is reducing cache usage, meaning that the algorithm has to fallback to slower caches less in the \texttt{int8} case.
However, this advantage is significantly reduced when we tune.
We believe that Ansor is not fully exploiting the performance potential and search space of \texttt{int8}, since it was
\begin{enumerate*}
    \item initially designed with \texttt{float32} computation in mind, and
    \item \texttt{int8} can require a more complex sequence of instructions to be generated to run efficiently, which Ansor does not appear to factor in.
\end{enumerate*}

For \texttt{float16} on the GPUs, contrasting to the CPUs, we observed in general a reduction in inference time on both GPUs, as they have direct hardware support for \texttt{float16} instructions.
Like the CPU, we also saw speedups with \texttt{int8} models, in general marginally higher than with \texttt{float16}.
However, if we take the best times across all algorithms, we did not see any cases where \texttt{float16} or \texttt{int8} achieved close to an ideal speedup of 2$\times$ and 4$\times$ relative to \texttt{float32}.

\subsection{Algorithms \& Data Formats}

We made several observations for algorithms and how they interacted with other layers of DLAS.
In the dense case, when tuned, \textit{VII) spatial-pack convolution is generally the best algorithm on the CPU (when tuned)}, and \textit{VIII) direct convolution is generally the best algorithm on the GPU (when tuned).}
Observation \textit{VIII)} goes against conventional wisdom, where we would normally expect \emph{GEMM} to be faster on the GPU.
However, we should note that we are using a custom implementation of GEMM within TVM, rather than an optimized BLAS library.
When the algorithm was not tuned, the best algorithm varied more, especially on the GPUs.
For example, with ImageNet on the HiKey, \emph{GEMM} was faster for smaller models, and \emph{direct} was faster for larger models.
In the sparse case, \textit{IX) GEMM is generally the best algorithm on CPUs, and direct is generally the best on GPU.}
For quantized models, the best algorithm varied on the CPUs and HiKey GPU, whereas on the Xavier \emph{direct} was generally the fastest.

In the evaluation we used the same convolution algorithm for all layers of a given model, but we could vary the algorithm used per-layer, which could bring significant performance speedups.
However, as other work has noted, data format transformation overheads between layers need to be considered~\cite{andersonOptimalDNNPrimitive2018,pradoLearningInferRLbased2019}.
We also did not explore all possible algorithms available, such as Winograd convolution~\cite{lavinFastAlgorithmsConvolutional2016}.
However, we chose three common algorithms to keep the across-stack evaluation tractable. 
Other works have explored algorithmic trade-offs in more detail~\cite{dolzPerformanceEnergyTradeoffs2023,congMinimizingComputationConvolutional2014,wenTASOTimeSpace2020,rovderOptimisingConvolutionalNeural2019}.

As discussed in Section~\ref{subsec:discuss:pruning}, \textit{X) for the pruning techniques we rarely observed the expected performance improvements} when compared to the dense implementations.
Partly this can be attributed to the inherent overheads of sparse data formats - CSR must store up to 3 values for every non-zero element.
This, coupled with irregular data access patterns means that the sparse algorithms do not realize their full potential.
CSR is not the only way to represent sparsity, and alternative data formats (and their complementary algorithms) may provide different trade-offs~\cite{langrEvaluationCriteriaSparse2016,gustavsonTwoFastAlgorithms1978,cipollettaEfficiencySparseTiledTensor2021}.
For example, for channel pruning we could store a list of the indices of pruned channels, and store the non-pruned channel data in a dense format.
This could allow us to leverage a more `dense-like' algorithm, with an overhead for looking up pruned indices.
Alternatively, we could use a format called `block-sparse row' (BSR), which is similar to CSR but represents blocks of sparse parameters, rather than individual weights.
This could allow us to reduce the overheads of the sparse storage format, with greater savings with larger block sizes at the risk or higher accuracy loss.
However, to fully exploit this we would need to change the pruning method to prune blocks.

\subsection{Systems Software}
\label{subsec:discuss:sys_software}

\textit{XI) auto-tuned code dramatically accelerates inference time, and can change the best algorithm or compression technique.}
However, we observed no speedup tuning the Xavier.
When testing with server-class Nvidia GPU platforms available to us, we observed speedups using auto-scheduling, using the same evaluation code.
Our conclusion is that some aspect of the AGX Xavier's software stack was incompatible with Ansor, but not in a way we could detect.

It is well known that auto-scheduling can provide significant speedups~\cite{zhengAnsorGeneratingHighPerformance2020,zhengDietCodeAutomaticOptimization2022}, however the search time required is non-negligible.
We could not collect auto-scheduled results on the HiKey for the ImageNet models, since they took too long to tune (over 140 hours each).
This highlights a key issue with auto-tuning, namely the cost of search, especially on constrained devices.
Approaches which significantly prune the search space~\cite{tollenaereAutotuningConvolutionsEasier2023} and techniques such as transfer-tuning~\cite{gibsonTransferTuningReusingAutoSchedules2023} can reduce the search time.

As highlighted in Section~\ref{subsubsec:discuss_quant}, we observed that \textit{XII) Pruned models saw a lower relative speedup when tuned}.
A more specialized sparse compiler, such as the emerging SparseTIR system~\cite{yeSparseTIRComposableAbstractions2023} could reduce the impact of these overheads.
This was also observed for the quantized models, which may require similar optimization support.

We initially experienced an issue compiling \texttt{int8} EfficientNet, as TVM assumed that in a multiplication operation only the left-hand operand would be pre-quantized.
However, the structure of EfficientNet violated this assumption, which necessitated a bug-fix which we pushed upstream.
This highlights that assumptions that systems software make about the properties of workloads may not always hold, especially when novel DNN architectures emerge.

All of the models were defined in PyTorch and evaluated in TVM, however there are other DNN frameworks available and the relative performance of different DNN frameworks has been well studied~\cite{hadidiCharacterizingDeploymentDeep2019,gibsonOptimizingGroupedConvolutions2020,louloudakisAssessingRobustnessImage2022}.
Other systems-software dimensions to consider are hand-tuned kernel libraries such as oneDNN~\cite{onednnOneDNN2020}, cuDNN~\cite{chetlurCuDNNEfficientPrimitives2014}, or other deep learning compilers such as TensorRT~\cite{nvidiacorporationNVIDIATensorRTProgrammable2016} or IREE~\cite{theireeauthorsIREE2019}.
Collage~\cite{jeonCollageSeamlessIntegration2023} can explore varying the backend for different subgraphs of the same DNN, which can bring significant performance improvements.

\subsection{Hardware}
\label{subsec:discuss:hardware}

As expected, \textit{XIII) the i7 CPU was generally faster than the HiKey CPU, and the Xavier GPU was generally faster than the HiKey GPU}.
Unfortunately we did not see improvements when tuning on the Xavier GPU.
Other aspects of the hardware that could be better utilized include the big.LITTLE architecture of the HiKey CPU (since we only leveraged the big cores), hyper-threading on Intel CPU (since we ran one thread per core), or leveraging both the CPU and GPU in parallel for the Xavier and HiKey.
However, across-stack optimizations would be required to exploit these features properly~\cite{wangHighThroughputCNNInference2020,loukadakisAcceleratingDeepNeural2018}.
We also did not leverage the Xavier GPU's 64 tensor cores in addition to its 512 general purpose CUDA cores.
Though TVM supports tensor cores, it requires manual schedule re-design for each algorithm.
MetaSchedule~\cite{shaoTensorProgramOptimization2022} can expand auto-scheduler search spaces to include hardware features such as tensor cores, which could allow us to more easily investigate this dimension.
As well as DNN focused extensions to general purpose hardware, we are also increasingly seeing DNN specific hardware accelerators.
These include fixed function accelerators such as the TPU~\cite{jouppiInDatacenterPerformanceAnalysis2017} and others~\cite{harisSECDATFLiteToolkitEfficient2023}; as well as reconfigurable accelerators such as MAERI~\cite{kwonMAERIEnablingFlexible2018} and others~\cite{munoz-martinezSTONNEEnablingCycleLevel2021,stjerngrenBifrostEndtoEndEvaluation2022,chenEyerissV2Flexible2019}.

\subsection{Evaluation Methodology}

For the experiments, we kept the batch size as 1, took the median of 150 runs, disregarding the first warm-up run.
Although this is a common deployment and evaluation scenario, it is important to be aware that this is not the only one, and experimental design should reflect which deployment case is being considered when evaluating models~\cite{wuMachineLearningFacebook2019,tangOptimizingGoogleWarehouse2013}.
For instance, for edge deployment we may expect the batch size to be small, whereas on the cloud it may be large.
Increased batch sizes mean increased memory requirements and inference latency, but also potentially higher throughput.
For the use of 150 runs, disregarding the first run, there could be deployment scenarios where we are more interested in the performance of these initial warm-up runs, before the cache behavior becomes more regular.

\section{Conclusions}

This paper first motivates and introduces the Deep Learning Acceleration Stack (DLAS) and then presents a perturbation study with an exploration of the impact of varying a small number of parameters at each layer of the stack.
In our study, we find a variety of across-stack interactions and scenarios where theoretical performance improvements were not achieved due to lack of full exploitation across the stack.
Our work is not intended to propose solutions to all of these limitations, instead highlights some complexities which emerge in deep learning acceleration and presents a conceptual framework (DLAS) for practitioners to approach their studies in the future.
We believe this can be achieved through closer collaboration across the layers of DLAS to enable more holistic co-design and co-optimization.

\bibliographystyle{ACM-Reference-Format}
\bibliography{99-references}

\end{document}